\documentclass{article}

 \usepackage[main, final]{neurips_2026}

\usepackage[utf8]{inputenc} % allow utf-8 input
\usepackage[T1]{fontenc}    % use 8-bit T1 fonts
\usepackage{hyperref}       % hyperlinks
\usepackage{url}            % simple URL typesetting
\usepackage{booktabs}       % professional-quality tables
\usepackage{amsfonts}       % blackboard math symbols
\usepackage{nicefrac}       % compact symbols for 1/2, etc.
\usepackage{microtype}      % microtypography
\usepackage{custom_commands}
\usepackage{wrapfig}

\title{LESSViT: Robust Hyperspectral Representation Learning under Spectral Configuration Shift}

\author{%
  Haozhe Si\textsuperscript{1}, 
  Yuxuan Wan\textsuperscript{1}, 
  Yuqing Wang\textsuperscript{1}, 
  Minh N. Do\textsuperscript{1,2}, 
  Han Zhao\textsuperscript{1}\\
  \textsuperscript{1}University of Illinois Urbana-Champaign, IL, USA\\
  \textsuperscript{2}VinUniversity, Hanoi, Vietnam \\
}

\begin{document}

\maketitle

\begin{abstract} 
Modeling hyperspectral imagery (HSI) across different sensors presents a fundamental challenge due to variations in wavelength coverage, band sampling, and channel dimensionality. As a result, models trained under a fixed spectral configuration often fail to generalize to other sensors. Existing Vision Transformer (ViT) approaches either rely on implicit spectral modeling with fixed channel assumptions or adopt explicit spatial–spectral attention with prohibitive computational cost, leading to a fundamental trade-off between efficiency and expressiveness. In this work, we introduce Low-rank Efficient Spatial–Spectral ViT (LESSViT), a sensor-flexible architecture for cross-spectral generalization. LESSViT is built on LESS Attention, a structured low-rank factorization that models joint spatial–spectral interactions through separable spatial and spectral components, reducing the complexity of full spatial–spectral attention from $\mathcal{O}(N^2 C^2)$ to $\mathcal{O}(rNC)$, where $N$ is the number of spatial tokens, $C$ is the number of spectral channels, and $r$ is the rank of the low-rank approximation. We further incorporate channel-agnostic patch embedding and wavelength-aware positional encoding to support flexible spectral inputs. To enable efficient and robust pretraining, we introduce a hyperspectral masked autoencoder (HyperMAE) with decoupled spatial–spectral masking and hierarchical channel sampling. We evaluate LESSViT under a cross-spectral generalization setting that simulates cross-sensor variability. Experiments on the SpectralEarth benchmark demonstrate that LESSViT improves robustness under spectral shifts while remaining competitive in-distribution, and explicit and efficient spatial–spectral modeling is essential for scalable and generalizable hyperspectral representation learning. The code is available through the project page: \texttt{https://uiuctml.github.io/LESSViT/}.
\end{abstract}

\begin{figure}
    \centering
    \includegraphics[width=\linewidth]{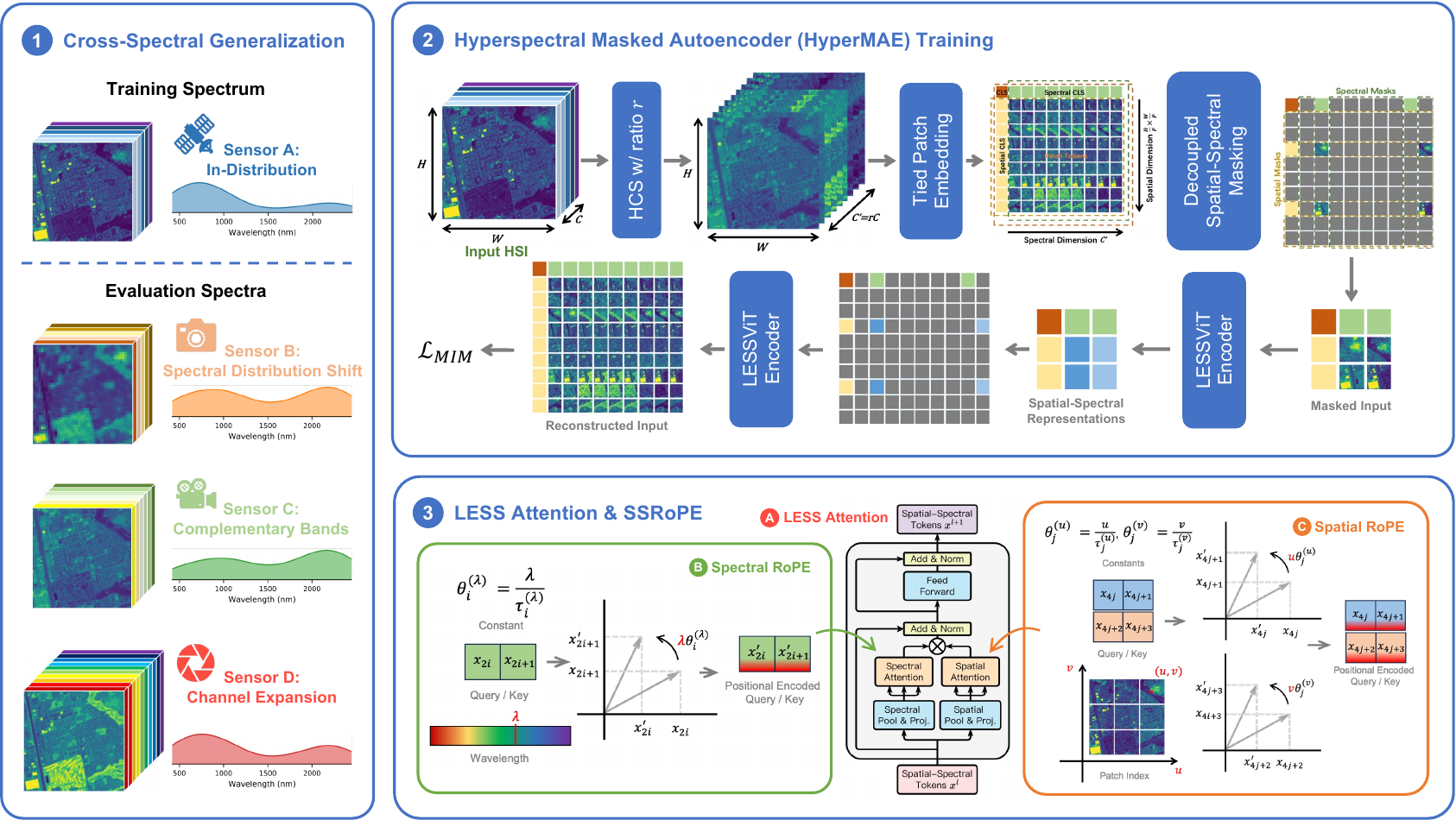}
    \caption{\textbf{Overview of LESSViT for cross-spectral generalization.}
    \textcircled{\scriptsize 1}: Cross-spectral generalization setting: train on a fixed spectral configuration and evaluate across spectral configurations with varying wavelength coverage and channel configurations.
    \textcircled{\scriptsize 2}: HyperMAE pretraining with decoupled spatial–spectral masking and hierarchical channel sampling for scalable and robust learning.
    \textcircled{\scriptsize 3}: LESS Attention with SSRoPE for efficient spatial–spectral modeling.
    }
    \label{fig:overview}
\end{figure}

\section{Introduction}\label{sec:intro}
Hyperspectral imagery (HSI) captures fine-grained spectral signatures across hundreds of contiguous wavelength bands, enabling detailed characterization of material composition~\citep{waste, counterfeit, medical, astronomy}, biophysical states~\citep{agriculture}, and environmental processes~\citep{climate, forest_hsi}. Unlike natural images, where channels correspond to broad RGB responses, hyperspectral channels represent physically grounded measurements indexed by wavelength. In practice, different platforms exhibit substantial differences in wavelength coverage, band sampling, and channel dimensionality~\citep{enmap, EO1H, specimiq}. Consequently, the same scene can produce sensor-dependent spectral responses, introducing a spectral shift that challenges generalization across configurations.

A key challenge in hyperspectral modeling is achieving generalization across varying spectral configurations. Unlike natural image domains with fixed RGB channels, hyperspectral imagery exhibits substantial variation across sensors in wavelength coverage, band sampling, and channel dimensionality, causing models trained under a fixed spectral configuration to fail under varying channel configurations. A natural approach is to explicitly model joint spatial–spectral structure, where spatial variation captures material distribution and spectral variation encodes wavelength-dependent responses. For example, crop-type recognition requires jointly interpreting field structure and vegetation spectra~\citep{zhao2020robust}, while flood mapping combines spatial context with water-specific spectral absorption~\citep{mateo2021towards}. However, such explicit modeling in Vision Transformers~\citep{vit} incurs $\mathcal{O}(N^2 C^2)$ complexity when treating channels as independent tokens, where $N$ is the number of spatial tokens and $C$ is the number of spectral channels, making it computationally prohibitive for hyperspectral inputs. This leads to a fundamental trade-off between efficient but implicit spectral modeling and expressive but computationally expensive spatial–spectral attention.

Existing approaches fail to fully resolve this trade-off. In practice, Vision Transformers are often adapted to hyperspectral data by collapsing the spectral dimension into the channel dimension~\citep{spectralearth, satmae}, where cross-channel interactions are only implicitly modeled within the initial patch embedding layer. While computationally efficient, this design assumes a fixed channel structure and treats spectral channels as unordered features, limiting its ability to capture wavelength-dependent relationships and to generalize across varying spectral configurations. DOFA~\citep{dofa} partially alleviates this limitation by introducing wavelength-conditioned patch embeddings, but does not explicitly model spatial–spectral interactions beyond the embedding stage. A more expressive alternative is to treat spectral channels as independent tokens and apply full attention over the resulting spatial–spectral grid~\citep{channel_vit}. This enables explicit spatial–spectral modeling, but incurs $\mathcal{O}(N^2 C^2)$ complexity, making it computationally prohibitive for hyperspectral inputs with hundreds of channels. HyperSigma~\citep{hypersigma} reduces this cost via decoupled spatial and spectral branches, but limits direct interaction between them, restricting its ability to capture fine-grained joint dependencies.

To address this fundamental trade-off, we introduce LESS Attention, a structured low-rank factorization of spatial–spectral attention. We exploit the structured interaction between spatial and spectral variations to decompose joint interactions into spatial-only and spectral-only components, while preserving their coupling through a low-rank tensor decomposition. This enables explicit modeling of joint spatial–spectral dependencies without computing full attention over the entire spatial–spectral grid, reducing complexity from $\mathcal{O}(N^2 C^2)$ to $\mathcal{O}(rNC)$, where $r$ controls the approximation rank.

Building on LESS Attention, we develop Low-rank Efficient Spatial–Spectral Vision Transformer (LESSViT), a scalable architecture for cross-spectral generalization. LESSViT employs a tied patch embedding for channel-agnostic tokenization and extends RoPE~\citep{rope} to jointly encode spatial coordinates and continuous wavelength information. Together, these designs ensure consistent spatial–spectral structure under varying spectral configurations. To enable scalable training of LESSViT, we adopt a hyperspectral masked autoencoder (HyperMAE) paradigm with decoupled spatial–spectral masking and hierarchical channel sampling, making pretraining both tractable under high channel dimensionality and robust to varying spectral configurations. Together, these designs enable explicit, efficient, and flexible spatial–spectral modeling across varying spectral regimes, as summarized in \Cref{fig:overview}.

We evaluate LESSViT on the SpectralEarth benchmark under a cross-spectral generalization setting that simulates cross-sensor spectral variation. Models are trained on a fixed channel set and evaluated across varying spectral configurations to reflect realistic sensor variability. Results show that LESSViT consistently improves robustness under spectral shifts while remaining competitive under in-distribution settings and scaling effectively to high channel counts. These findings highlight that explicit and efficient spatial–spectral modeling is essential for generalizable hyperspectral representation learning. Our contributions are threefold:
\begin{itemize}
\item We propose LESS Attention, a structured low-rank factorization of spatial–spectral attention that reduces the complexity of modeling joint spatial–spectral interactions.
\item We develop LESSViT, a channel-agnostic spatial–spectral Vision Transformer with wavelength-aware positional encoding, enabling consistent and robust modeling under varying spectral configurations.
\item We establish a cross-spectral generalization evaluation protocol on the SpectralEarth benchmark to systematically assess cross-spectral generalization, demonstrating the effectiveness of explicit and scalable spatial–spectral modeling.
\end{itemize}
% \han{We never refer to Fig. 1 in the introduction, but it provides a nice overview of the method. We should refer to it in the text.}
\section{Low-rank Efficient Spatial-Spectral ViT (LESSViT)}
To achieve generalization across spectral configurations, a model must satisfy three key requirements: \textit{i)}~Channel-agnostic tokenization, to support varying spectral dimensionality; \textit{ii)}~Explicit spatial–spectral modeling, to capture wavelength-dependent structure; and \textit{iii)}~Computational scalability, to remain tractable for high-dimensional inputs.
Existing architectures fail to satisfy these requirements simultaneously. To address these challenges, we propose LESSViT, a spatial–spectral ViT that integrates structured tokenization and efficient attention to enable scalable and robust modeling under varying spectral configurations.

\subsection{Tied Patch Embedding} \label{sec:embedding}
A key requirement for cross-spectral generalization is that tokenization should not depend on a fixed channel structure. Standard patch embedding layers assume a fixed number of input channels, making them incompatible with varying spectral configurations. To address this limitation, we employ a \emph{tied patch embedding} layer~\citep{channel_vit} with shared projection weights across channels. Given an input hyperspectral image of size $C \times H \times W$, we partition it into $C \times \frac{H}{P} \times \frac{W}{P}$ patches of size $P \times P$. Each patch is projected into a $D$-dimensional token using a shared linear mapping $W \in \mathbb{R}^{P^2 \times D}$. This weight-sharing mechanism~\citep{shared_pe} ensures channel-agnostic processing while preserving per-channel structure, enabling the model to operate across varying spectral dimensionalities. The resulting tokens form a spatial–spectral grid of size $\mathbb{R}^{N \times C \times D}$, where $N = \frac{H}{P} \times \frac{W}{P}$. This representation retains the structured organization of hyperspectral data and serves as the foundation for subsequent spatial–spectral modeling.

\begin{figure}[t]
\centering
\subfloat[\textbf{Spatial-Spectral Tokenization.}]
{%
\centering
\includegraphics[width=0.33\columnwidth]{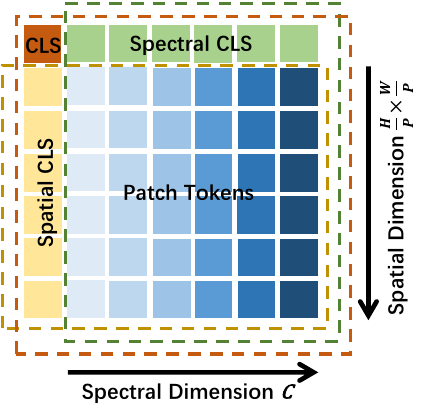}
\label{fig:hpe}
}
\hfill
\subfloat[\textbf{LESS Attention Block.}]{%
\includegraphics[width=0.56\columnwidth]{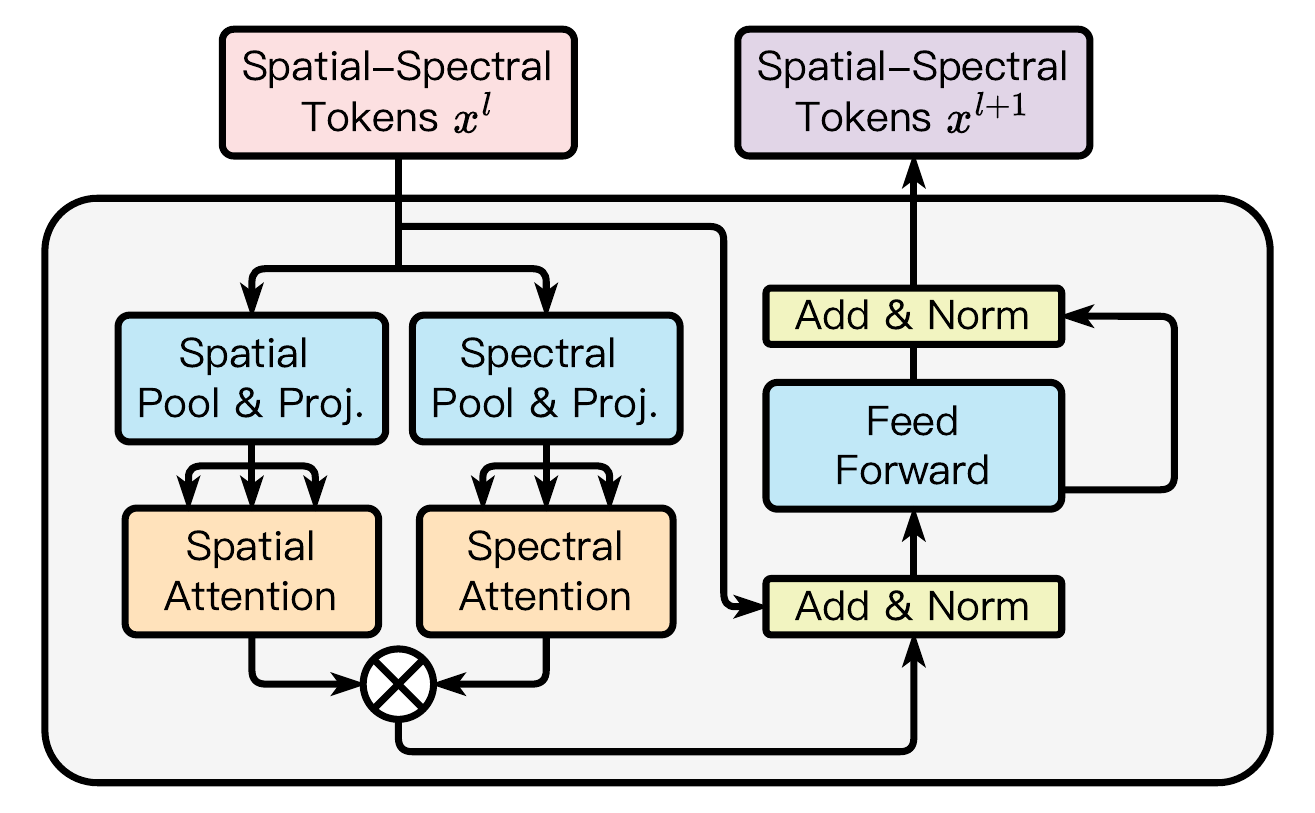}
\label{fig:less-block}
}
\caption{\textbf{Overview of LESSViT.}
(a) The tied patch embedding converts a hyperspectral image (\(C\times H\times W\)) into a grid of spatial–spectral tokens with spatial, spectral, and global \texttt{[CLS]} tokens.
(b) The LESS block factorizes spatial and spectral attention via structured decomposition, enabling efficient modeling of joint spatial–spectral interactions.}
\label{fig:module}
\end{figure}

As illustrated in \Cref{fig:hpe}, we incorporate learnable spatial, spectral, and global \texttt{[CLS]} tokens to facilitate structured aggregation. Spatial \texttt{[CLS]} tokens summarize information along the spectral dimension, spectral \texttt{[CLS]} tokens summarize information along the spatial dimension, and a global \texttt{[CLS]} token captures joint spatial–spectral information. This yields an augmented token grid of size $\mathbb{R}^{(N+1)\times(C+1)\times D}$, providing structured summaries for the spatial-spectral token grids.

\subsection{Low-rank Efficient Spatial-Spectral Attention}\label{sec:less}
To model interactions over spatial–spectral tokens $X \in \mathbb{R}^{N \times C \times D}$, a direct approach flattens tokens and applies standard self-attention~\citep{channel_vit}. However, computing self-attention for $NC$ tokens incurs $\mathcal{O}(N^2 C^2 D)$ complexity, which is prohibitive for hyperspectral inputs. Consequently, explicit spatial–spectral modeling in this form becomes computationally intractable, failing to meet the scalability requirement. To address this limitation, we exploit the structural separability of hyperspectral data: spatial and spectral variations capture complementary factors and admit a low-rank decomposition. This enables joint spatial–spectral interactions to be modeled through separable components while preserving their coupling. 

Based on this observation, we propose the Low-rank Efficient Spatial-Spectral (LESS) attention block, which models joint spatial–spectral attention through a structured low-rank decomposition. Specifically, LESS Attention decomposes interactions into spatial-only and spectral-only components, whose coupling is captured through a low-rank composition. For simplicity, the following formulation considers a single attention head.  As shown in Algorithm~\ref{algo:low-rank-spatial-spectral}, our LESS attention block first decomposes input tokens $X$ into spatial tokens $X_S$ and spectral tokens $X_C$ using an Attention Pooling Layer (\textsc{AttenPool}). \textsc{AttenPool} aggregates tokens along a selected dimension by using the corresponding \texttt{[CLS]} token as a query, producing a weighted summary of all tokens that is then linearly projected to the target sub-dimension.

We then compute spatial attention $A_S$ and spectral attention $A_C$ independently using $X_S$ and $X_C$. Their Kronecker product defines a structured low-rank form of joint spatial–spectral attention:
 % \han{here we are not approximating the attention matrix, but rather directly modeling it as a low-rank decomposition. Let's use defeq to make this clear.}
\begin{equation}
A \coloneqq \sum_{i=1}^{r} A_C^i \otimes A_S^i.
 \end{equation}
Accordingly, the attention is given by 
% \han{similar comment here about the output being directly modeled as a low-rank decomposition, rather than an approximation of the full attention output.}
\begin{equation}
Y \coloneqq \sum_{i=1}^{r} (A_C^i \otimes A_S^i)(V_C \otimes V_S)
= \sum_{i=1}^{r} (A_C^i V_C) \otimes (A_S^i V_S) = \sum_{i=1}^{r} Y_C^i \otimes Y_S^i,
\end{equation}
where $V_S \in \mathbb{R}^{N \times d_s}$ and
$V_C \in \mathbb{R}^{C \times d_c}$ are the spatial and spectral
value representations, respectively, and
$Y_S^i \in \mathbb{R}^{N \times d_s}$ and
$Y_C^i \in \mathbb{R}^{C \times d_c}$ are the corresponding branch outputs. This formulation avoids constructing the full $NC \times NC$ attention matrix and instead computes interactions through separable spatial and spectral components.

\setlength{\textfloatsep}{5pt}
\begin{algorithm*}[t]
    \caption{Low-Rank Efficient Spatial-Spectral (LESS) Attention Block}\label{algo:low-rank-spatial-spectral}
    \begin{algorithmic}[1]
        \Require Input tokens $X \in \mathbb{R}^{N \times C \times D}$, Sub-dimensions $d_sd_c=D$
        \Ensure Output tokens $Y \in \mathbb{R}^{N \times C \times D}$
        \State \textbf{\# Decompose spatial-spectral tokens}
        \State $X_S \gets \textsc{AttenPool}(X, \text{dim}=1)\in \mathbb{R}^{N \times d_s}$ \Comment{Spatial-only tokens}
        \State $X_C \gets \textsc{AttenPool}(X, \text{dim}=0)\in \mathbb{R}^{C \times d_c}$ \Comment{Spectral-only tokens}
        \State \textbf{\# Compute spatial-only attention}
        \State $Q_S, K_S, V_S \gets X_S W_{Q_S}, X_S W_{K_S}, X_S W_{V_S}\in \mathbb{R}^{N \times d_s}$
        \State $Y_S \gets \textsc{Softmax}(Q_S K_S^\top/\sqrt{d_s})V_S\in \mathbb{R}^{N \times d_s}$ \Comment{Complexity $O(N^2d_s)$}
        \State \textbf{\# Compute spectral-only attention}
        \State $Q_C, K_C, V_C \gets X_C W_{Q_C}, X_C W_{K_C}, X_C W_{V_C}\in \mathbb{R}^{C \times d_c}$
        \State $Y_C \gets \textsc{Softmax}(Q_C K_C^\top/\sqrt{d_c})V_C\in \mathbb{R}^{C \times d_c}$ \Comment{Complexity $O(C^2d_c)$}
        \State \textbf{\# Approximate the spatial-spectral attention}
        \State $Y \gets Y_C\otimes Y_S\in \mathbb{R}^{NC \times d_sd_c}$ \Comment{Complexity $O(NCD)$}
        \State $Y \gets \textsc{Reshape}(Y, (N, C, D))$
        \State \Return $Y$
    \end{algorithmic}
\end{algorithm*}

The resulting complexity is reduced from $\mathcal{O}(N^2 C^2 D)$ to $\mathcal{O}(rN^2 d_s + rC^2 d_c + rNCD)$, where $d_s d_c = D$ and $r \ll \min(N, C)$. Under the regime where spatial and spectral token counts are balanced relative to the sub-dimensions (i.e., $N/C < d_c$ and $C/N < d_s$), the mixed term $NCD$ dominates the overall complexity. As a result, LESS Attention enables practical scaling to hyperspectral inputs with hundreds of channels, where full spatial–spectral attention is otherwise infeasible.

In practice, we allocate a larger sub-dimension $d_s$ to spatial features and a smaller $d_c$ to spectral features, as spectral dependencies often exhibit lower intrinsic dimensionality. This choice preserves the desired complexity regime while maintaining modeling flexibility. Consequently, LESS Attention enables an efficient approximation of explicit spatial–spectral interactions, reconciling expressive joint modeling with computational scalability.

\textbf{Spatial-Spectral RoPE.}
To model spectral relationships consistently under varying spectral configurations, positional encoding must be independent of fixed channel indices and instead reflect the physical structure of the spectrum. We introduce spatial–spectral rotary positional embedding (SSRoPE), an axial design that applies standard 2D RoPE~\citep{2drope} over spatial coordinates $(u, v)$ and 1D RoPE~\citep{rope} over the spectral dimension using wavelengths $\lambda$. This yields relative positional encoding across both axes and aligns naturally with the factorized structure of LESS Attention, enabling separate spatial and spectral modeling while preserving their coupling. The detailed formulation is provided in \Cref{appendix:ssrope}.
% \han{To me this paragraph is not very clear. Can we give the precise mathematical formulation of SSRoPE and how it is applied to the tokens?}

\subsection{Hyperspectral Masked Autoencoder (HyperMAE)} \label{sec:multi-mae}
Applying standard masked autoencoding (MAE) to LESSViT is both structurally incompatible and inefficient under high channel dimensionality. We therefore introduce HyperMAE, a tailored pretraining framework for self-supervision.

\textbf{Decoupled spatial–spectral masking.}
Standard MAE applies random masking over flattened tokens, implicitly assuming a unified token space. This breaks the axis-aligned structure required by LESS Attention. Inspired by tube masking in VideoMAE~\citep{videomae}, we apply identical spatial masks across spectral channels rather than temporal frames, and independently mask spectral channels. We use a masking ratio of 75\% along each axis. This preserves the separable token grid required by the factorized attention while exposing the model to partial spatial and spectral observations.

\textbf{Hierarchical Channel Sampling (HCS).}
While LESS Attention enables efficient spatial–spectral modeling, the HyperMAE decoder for dense reconstruction increases memory and computation, especially at high channel counts. To mitigate this, we adopt hierarchical channel sampling~\citep{channel_vit}, where only a subset of channels is randomly sampled at each iteration within a predefined ratio range $[r_l, r_h]$. This reduces reconstruction cost while maintaining exposure to diverse partial-channel inputs. HCS is used only during pretraining. During fine-tuning and inference, the decoder is removed and replaced with a lightweight task head, and computation is dominated by the encoder. This allows efficient processing of full-channel inputs at test time without incurring dense reconstruction cost.
\section{Experiments}
We evaluate LESSViT under a cross-spectral generalization setting that captures sensor-induced spectral variability. Models are pretrained and fine-tuned on a fixed spectral configuration and evaluated under different channel configurations designed to isolate distinct forms of spectral shift. We first introduce the cross-spectral generalization task and the experimental protocol. We then present quantitative results and ablation studies to analyze performance, robustness, and efficiency.

\begin{figure}
    \centering
    \includegraphics[width=\linewidth]{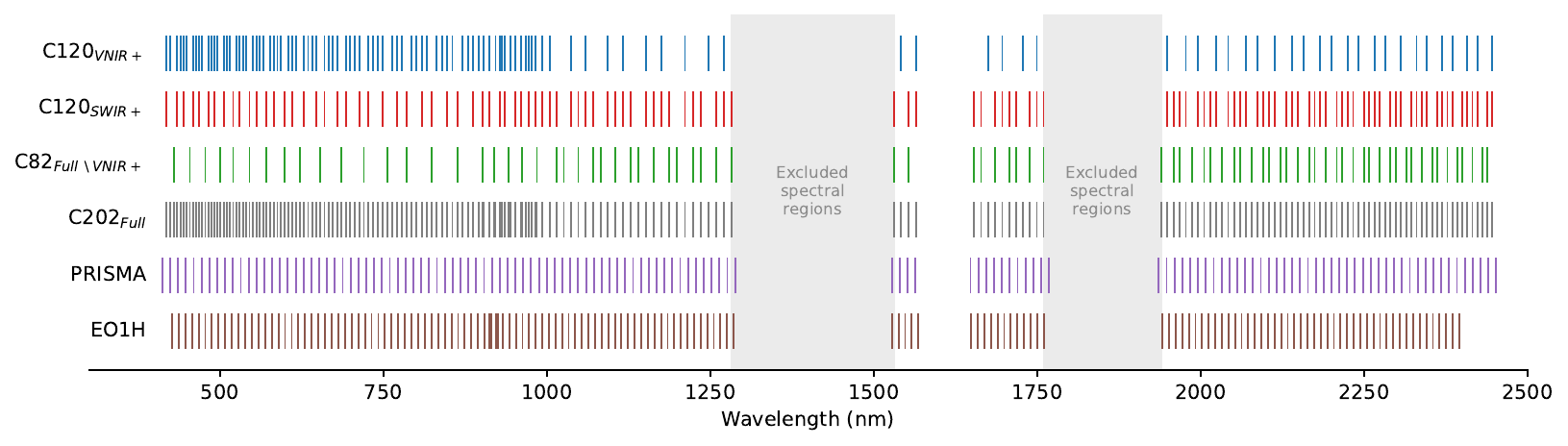}
    \caption{\textbf{Band-center locations of the six evaluation configurations.} Each tick marks a retained band center. $C120_{\text{VNIR+}}$ and $C120_{\text{SWIR+}}$ differ in their VNIR--SWIR sampling balance; $C82$ contains the complementary bands interleaved with the training bands, and $C202$ includes all available EnMAP bands. PRISMA and EO1H show the target band centers of the SRF-based configurations. Shaded regions indicate excluded spectral regions.}
    \label{fig:spectrum}
\end{figure}

\subsection{Cross-Spectral Generalization}\label{sec:benchmark}

SpectralEarth~\citep{spectralearth} provides a large-scale hyperspectral pretraining dataset and evaluation benchmark spanning diverse geospatial tasks. In practice, different sensors capture distinct spectral signatures due to variations in wavelength coverage and band sampling. Building on this dataset, we construct a cross-spectral generalization setting by varying channel configurations while keeping spatial content fixed, enabling evaluation that isolates spectral variation.

SpectralEarth provides 202 hyperspectral channels from EnMAP~\citep{enmap}. We partition the spectrum into visible and near-infrared (VNIR) and shortwave infrared (SWIR) regions using a threshold at 1000 nm, resulting in 100 VNIR channels and 102 SWIR channels. Based on this partition, we construct multiple channel subsets that emulate different sensor designs:

% \begin{itemize}
\textbf{Pro-VNIR configuration}~($C120_{\text{VNIR+}}$): 120 channels composed of 80 VNIR and 40 SWIR bands, uniformly sampled within each spectral region.

\textbf{Pro-SWIR configuration}~($C120_{\text{SWIR+}}$): 120 channels composed of 40 VNIR and 80 SWIR bands, evaluating robustness to a shift in spectral sampling from VNIR toward SWIR while keeping the channel count fixed.

\textbf{Complementary configuration}~($C82_{\text{Full} \setminus \text{VNIR+}}$): 82 channels comprising 20 VNIR and 62 SWIR bands that complement $C120_{\text{VNIR+}}$ and are spectrally interleaved with its bands, evaluating generalization to unobserved band centers through spectral interpolation.

\textbf{Full-spectrum configuration}~($C202_{\text{Full}}$): all available channels, representing a setting where
a model trained with 120 channels must generalize to a larger input channel set without retraining or architectural changes.
% \end{itemize}

\textbf{SRF-based configurations}: We synthesize PRISMA~\citep{prisma} and EO-1 Hyperion~\citep{EO1H} observations by convolving EnMAP spectra with Gaussian spectral response functions (SRFs) defined by each sensor's published band centers and full widths at half maximum (FWHMs). After excluding bands with insufficient spectral support or strong atmospheric absorption, we retain 133 PRISMA-like and 155 Hyperion-like bands. These settings test the model robustness to sensor-dependent spectral mixing beyond channel-subset selection.

The band-center locations of all six configurations are visualized in \Cref{fig:spectrum}. In our experiments, models are pretrained and fine-tuned on the $C120_{\text{VNIR+}}$ configuration and evaluated under five settings:
(i) \emph{In-distribution} ($C120_{\text{VNIR+}}$),
(ii) \emph{Spectral sampling shift} ($C120_{\text{SWIR+}}$),
(iii) \emph{Complementary bands} ($C82_{\text{Full} \setminus \text{VNIR+}}$),
(iv) \emph{Channel expansion} ($C202_{\text{Full}}$), (v) \emph{SRF shift} (PRISMA and EO1H).

\subsection{Experimental Protocol}
\textbf{LESSViT Configuration.}~
We adopt a ViT-Base-sized architecture with 12 encoder and 8 decoder LESS attention blocks, using embedding dimensions of 768 and 512, respectively. Each block employs 12 attention heads. The LESS attention uses rank $r=1$ with attention ratio $\frac{d_s}{d_c}=16$, where $d_s$ and $d_c$ denote the per-head spatial and spectral sub-dimensions. The patch size is set to 16. LESSViT is pretrained on the $C120_{\text{VNIR+}}$ configuration of SpectralEarth using the HyperMAE framework for 200 epochs. Additional details are provided in \Cref{appendix:pretrain}.

\textbf{Baselines.}~
We compare LESSViT with state-of-the-art ViT-based geospatial representation learning methods, including SpectralViT~\citep{spectralearth}, DOFA~\citep{dofa}, HyperSigma~\citep{hypersigma}, and HyperFree~\citep{hyperfree}. The released SpectralViT model is pretrained on the full $C202_{\text{Full}}$ spectrum, which would introduce information leakage. We therefore retrain SpectralViT on $C120_{\text{VNIR+}}$ using the same protocol and data as LESSViT to ensure a fair comparison under varying spectral configurations. In contrast, DOFA, HyperSigma, and HyperFree use pretrained weights from their respective datasets and are not exposed to SpectralEarth channel configurations, so we directly adopt their released models. All compared models are subsequently fine-tuned under an identical protocol on $C120_{\text{VNIR+}}$ for 20 epochs, with hyperparameters selected on a held-out validation set, and evaluated across the six spectral configurations. Additional details are provided in \Cref{appendix:evaluation}. % ChannelViT is included only for efficiency comparison due to its high computational cost, with results reported in \Cref{sec:eff}.

\begin{table}[t]
\centering
\small
\renewcommand{\arraystretch}{1.12}
\setlength{\tabcolsep}{3.2pt}
\caption{\textbf{Cross-spectral generalization across five downstream tasks.} Scores are means over five seeds; percentages indicate relative drops from the in-distribution configuration. \textbf{Bold} denotes the highest score within each dataset–configuration pair and the lowest average relative drop in the summary block.}
\label{tab:benchmark}
\resizebox{\textwidth}{!}{%
\begin{tabular}{lcccccccccccc}
\toprule
 & \multicolumn{6}{c}{CDL (mIoU)} & \multicolumn{6}{c}{EuroCrops (mIoU)} \\
\cmidrule(lr){2-7}\cmidrule(lr){8-13}
Model & $C120_{\mathrm{VN+}}$ & $C120_{\mathrm{SW+}}$ & $C82$ & $C202$ & PRISMA & EO1H & $C120_{\mathrm{VN+}}$ & $C120_{\mathrm{SW+}}$ & $C82$ & $C202$ & PRISMA & EO1H \\
\midrule
SpectralViT & $\mathbf{73.33}$ & $15.11_{\downarrow 79.4\%}$ & $12.36_{\downarrow 83.1\%}$ & $5.59_{\downarrow 92.4\%}$ & $10.84_{\downarrow 85.2\%}$ & $7.81_{\downarrow 89.4\%}$ & $\mathbf{71.69}$ & $26.27_{\downarrow 63.4\%}$ & $41.14_{\downarrow 42.6\%}$ & $46.28_{\downarrow 35.4\%}$ & $42.23_{\downarrow 41.1\%}$ & $29.32_{\downarrow 59.1\%}$ \\
HyperSigma & $44.48$ & $13.93_{\downarrow 68.7\%}$ & $14.50_{\downarrow 67.4\%}$ & $14.51_{\downarrow 67.4\%}$ & $10.08_{\downarrow 77.3\%}$ & $12.86_{\downarrow 71.1\%}$ & $56.54$ & $52.16_{\downarrow 7.7\%}$ & $48.45_{\downarrow 14.3\%}$ & $51.59_{\downarrow 8.8\%}$ & $46.32_{\downarrow 18.1\%}$ & $38.84_{\downarrow 31.3\%}$ \\
DOFA & $40.06$ & $18.38_{\downarrow 54.1\%}$ & $18.08_{\downarrow 54.9\%}$ & $12.39_{\downarrow 69.1\%}$ & $13.06_{\downarrow 67.4\%}$ & $6.39_{\downarrow 84.1\%}$ & $55.59$ & $36.85_{\downarrow 33.7\%}$ & $32.83_{\downarrow 40.9\%}$ & $31.52_{\downarrow 43.3\%}$ & $50.28_{\downarrow 9.5\%}$ & $50.40_{\downarrow 9.3\%}$ \\
HyperFree & $42.50$ & $28.48_{\downarrow 33.0\%}$ & $16.22_{\downarrow 61.8\%}$ & $22.16_{\downarrow 47.9\%}$ & $28.09_{\downarrow 33.9\%}$ & $19.59_{\downarrow 53.9\%}$ & $52.22$ & $46.63_{\downarrow 10.7\%}$ & $32.87_{\downarrow 37.1\%}$ & $36.25_{\downarrow 30.6\%}$ & $45.71_{\downarrow 12.5\%}$ & $20.91_{\downarrow 60.0\%}$ \\
\rowcolor[HTML]{ECF4FF}
LESSViT & $53.44$ & $\mathbf{48.32}_{\downarrow 9.6\%}$ & $\mathbf{42.08}_{\downarrow 21.3\%}$ & $\mathbf{50.75}_{\downarrow 5.0\%}$ & $\mathbf{48.96}_{\downarrow 8.4\%}$ & $\mathbf{49.05}_{\downarrow 8.2\%}$ & $57.91$ & $\mathbf{55.69}_{\downarrow 3.8\%}$ & $\mathbf{55.47}_{\downarrow 4.2\%}$ & $\mathbf{57.81}_{\downarrow 0.2\%}$ & $\mathbf{57.63}_{\downarrow 0.5\%}$ & $\mathbf{57.48}_{\downarrow 0.7\%}$ \\
\bottomrule
\end{tabular}}
\par\vspace{6pt}
\resizebox{\textwidth}{!}{%
\begin{tabular}{lcccccccccccc}
\toprule
 & \multicolumn{6}{c}{BDFORET (mIoU)} & \multicolumn{6}{c}{BNETD (mIoU)} \\
\cmidrule(lr){2-7}\cmidrule(lr){8-13}
Model & $C120_{\mathrm{VN+}}$ & $C120_{\mathrm{SW+}}$ & $C82$ & $C202$ & PRISMA & EO1H & $C120_{\mathrm{VN+}}$ & $C120_{\mathrm{SW+}}$ & $C82$ & $C202$ & PRISMA & EO1H \\
\midrule
SpectralViT & $\mathbf{77.17}$ & $48.31_{\downarrow 37.4\%}$ & $48.68_{\downarrow 36.9\%}$ & $48.73_{\downarrow 36.9\%}$ & $46.77_{\downarrow 39.4\%}$ & $44.55_{\downarrow 42.3\%}$ & $\mathbf{44.69}$ & $16.86_{\downarrow 62.3\%}$ & $17.06_{\downarrow 61.8\%}$ & $16.66_{\downarrow 62.7\%}$ & $18.00_{\downarrow 59.7\%}$ & $16.07_{\downarrow 64.1\%}$ \\
HyperSigma & $63.76$ & $47.97_{\downarrow 24.8\%}$ & $44.15_{\downarrow 30.8\%}$ & $48.94_{\downarrow 23.2\%}$ & $43.68_{\downarrow 31.5\%}$ & $9.54_{\downarrow 85.0\%}$ & $30.36$ & $18.78_{\downarrow 38.1\%}$ & $17.64_{\downarrow 41.9\%}$ & $17.68_{\downarrow 41.8\%}$ & $15.31_{\downarrow 49.6\%}$ & $7.22_{\downarrow 76.2\%}$ \\
DOFA & $58.14$ & $51.26_{\downarrow 11.8\%}$ & $42.03_{\downarrow 27.7\%}$ & $50.45_{\downarrow 13.2\%}$ & $45.04_{\downarrow 22.5\%}$ & $2.36_{\downarrow 95.9\%}$ & $28.88$ & $18.06_{\downarrow 37.5\%}$ & $18.87_{\downarrow 34.7\%}$ & $18.89_{\downarrow 34.6\%}$ & $15.75_{\downarrow 45.5\%}$ & $6.29_{\downarrow 78.2\%}$ \\
HyperFree & $58.42$ & $54.43_{\downarrow 6.8\%}$ & $52.63_{\downarrow 9.9\%}$ & $54.13_{\downarrow 7.3\%}$ & $54.44_{\downarrow 6.8\%}$ & $53.93_{\downarrow 7.7\%}$ & $29.53$ & $18.53_{\downarrow 37.3\%}$ & $16.87_{\downarrow 42.9\%}$ & $21.47_{\downarrow 27.3\%}$ & $21.93_{\downarrow 25.7\%}$ & $18.97_{\downarrow 35.8\%}$ \\
\rowcolor[HTML]{ECF4FF}
LESSViT & $63.87$ & $\mathbf{62.43}_{\downarrow 2.3\%}$ & $\mathbf{60.58}_{\downarrow 5.2\%}$ & $\mathbf{63.35}_{\downarrow 0.8\%}$ & $\mathbf{62.03}_{\downarrow 2.9\%}$ & $\mathbf{62.60}_{\downarrow 2.0\%}$ & $32.37$ & $\mathbf{30.76}_{\downarrow 5.0\%}$ & $\mathbf{26.35}_{\downarrow 18.6\%}$ & $\mathbf{31.25}_{\downarrow 3.5\%}$ & $\mathbf{31.80}_{\downarrow 1.8\%}$ & $\mathbf{29.62}_{\downarrow 8.5\%}$ \\
\bottomrule
\end{tabular}}
\par\vspace{6pt}
\resizebox{0.8\textwidth}{!}{%
\begin{tabular}{lccccccccccc}
\toprule
 & \multicolumn{6}{c}{CORINE (mAP)} & \multicolumn{5}{c}{Avg. Rel. Drop (\%)} \\
\cmidrule(lr){2-7}\cmidrule(lr){8-12}
Model & $C120_{\mathrm{VN+}}$ & $C120_{\mathrm{SW+}}$ & $C82$ & $C202$ & PRISMA & EO1H & $C120_{\mathrm{SW+}}$ & $C82$ & $C202$ & PRISMA & EO1H \\
\midrule
SpectralViT & $\mathbf{78.93}$ & $36.90_{\downarrow 53.2\%}$ & $36.92_{\downarrow 53.2\%}$ & $38.37_{\downarrow 51.4\%}$ & $35.24_{\downarrow 55.4\%}$ & $33.08_{\downarrow 58.1\%}$ & 59.1 & 55.5 & 55.8 & 56.2 & 62.6 \\
HyperSigma & $69.21$ & $36.05_{\downarrow 47.9\%}$ & $27.56_{\downarrow 60.2\%}$ & $47.11_{\downarrow 31.9\%}$ & $38.60_{\downarrow 44.2\%}$ & $39.48_{\downarrow 43.0\%}$ & 37.4 & 42.9 & 34.6 & 44.1 & 61.3 \\
DOFA & $66.40$ & $37.89_{\downarrow 42.9\%}$ & $48.27_{\downarrow 27.3\%}$ & $35.33_{\downarrow 46.8\%}$ & $25.66_{\downarrow 61.4\%}$ & $17.58_{\downarrow 73.5\%}$ & 36.0 & 37.1 & 41.4 & 41.3 & 68.2 \\
HyperFree & $74.04$ & $47.93_{\downarrow 35.3\%}$ & $63.31_{\downarrow 14.5\%}$ & $68.48_{\downarrow 7.5\%}$ & $65.83_{\downarrow 11.1\%}$ & $62.28_{\downarrow 15.9\%}$ & 24.6 & 33.2 & 24.1 & 18.0 & 34.6 \\
\rowcolor[HTML]{ECF4FF}
LESSViT & $76.35$ & $\mathbf{74.21}_{\downarrow 2.8\%}$ & $\mathbf{72.95}_{\downarrow 4.5\%}$ & $\mathbf{75.99}_{\downarrow 0.5\%}$ & $\mathbf{75.05}_{\downarrow 1.7\%}$ & $\mathbf{75.01}_{\downarrow 1.8\%}$ & \textbf{4.7} & \textbf{10.7} & \textbf{2.0} & \textbf{3.0} & \textbf{4.2} \\
\bottomrule
\end{tabular}}
\end{table}
\subsection{Quantitative Results}\label{sec:quant_res}
\Cref{tab:benchmark} summarizes performance across six spectral configurations after fine-tuning on \(C120_{\mathrm{VN+}}\), with no further adaptation at evaluation. We report mIoU for segmentation and mAP for multi-label classification, averaged over five random seeds. Full results with standard deviations are provided in the appendix. For each task, the relative performance drop is computed as \(1-C_{\mathrm{OOD}}/C120_\mathrm{VN+}\). The summary block reports the unweighted average of these relative drops across the five tasks for each OOD configuration.

\textbf{Cross-spectral generalization.}
As shown in \Cref{tab:benchmark}, LESSViT achieves the highest mean performance across all 25 out-of-distribution task--configuration pairs, covering spectral sampling shifts, complementary bands, channel expansion, and simulated sensor responses. Its average relative performance drops range from 2.0\% to 10.7\%, compared with 18.0\%--34.6\% for HyperFree, the strongest baseline by this robustness measure. The complementary-band configuration ($C82$) produces the largest average drop for LESSViT (10.7\%), indicating that generalization to unobserved band centers remains challenging even when they are interleaved with training bands.
Under the simulated PRISMA and EO-1 Hyperion responses, LESSViT retains strong performance, with average drops of only 3.0\% and 4.2\%, respectively. These results demonstrate robustness to both channel-subset reconfiguration and simulated sensor-dependent spectral responses.

\textbf{In-distribution performance.}
Under the in-distribution setting ($C120_{\text{VNIR+}}$), SpectralViT consistently achieves the best performance, while LESSViT ranks second in most cases. The strong in-distribution performance suggests that SpectralViT is specialized to the fixed spectral configuration, enabling more effective exploitation of spatial patterns. 

Additionally, SpectralViT adopts a smaller patch size (4$\times$4), which preserves finer spatial details and benefits dense prediction tasks. Consistently, the performance gap is smaller on CORINE classification, where spatial resolution is less critical.

\textbf{Discussion.}
The observed results reveal a trade-off between in-distribution specialization and cross-spectral generalization. Models such as SpectralViT achieve strong performance under a fixed spectral configuration, but degrade substantially when the channel configuration or spectral response changes. These results suggest that strong in-distribution performance does not necessarily translate into robustness under spectral shifts. In contrast, LESSViT explicitly models spatial--spectral interactions and maintains more stable performance across complementary bands, channel expansion, and simulated SRF-induced spectral mixing. The PRISMA-like and Hyperion-like results extend this robustness beyond channel-subset selection to changes in how spectral measurements are formed. Overall, LESSViT achieves a more favorable balance between in-distribution performance and robustness across the evaluated spectral configurations.

\begin{figure}[t]
    \centering
    \includegraphics[width=\linewidth]{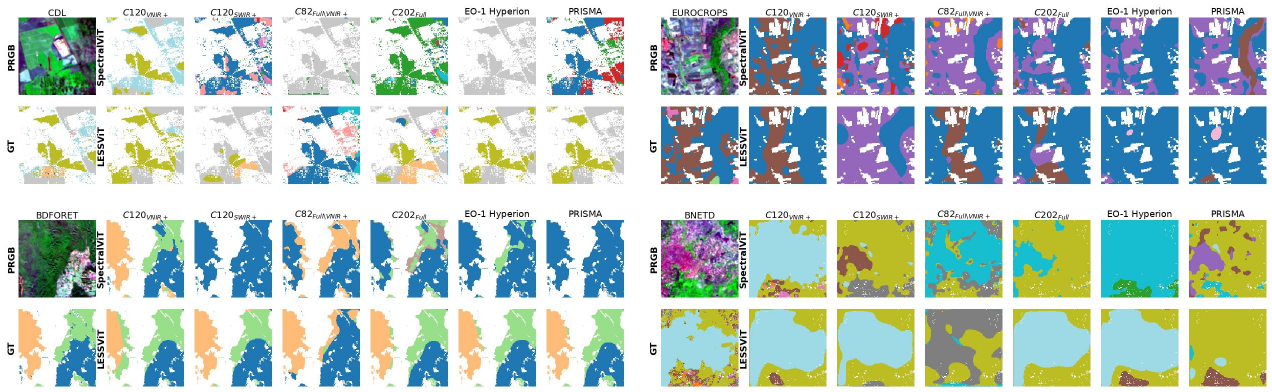}
    \caption{\textbf{Qualitative results under cross-spectral generalization.} PRGB denotes pseudo-RGB visualization of hyperspectral inputs, and GT denotes ground-truth segmentation masks. For clearer comparison with GT, background pixels are masked out in the predicted segmentation maps. We show results under different spectral configurations, including in-distribution, spectral shift, complementary bands, channel expansion, and simulated EO-1 Hyperion and PRISMA spectral responses.}
    \label{fig:vis}
\end{figure}

\subsection{Qualitative Results}
We visualize representative segmentation results under channel-subset reconfiguration in \Cref{fig:vis}, comparing the model with the strongest in-distribution performance (SpectralViT) and the model with the best robustness under varying spectral configurations (LESSViT). In the in-distribution setting, SpectralViT produces more detailed segmentation maps, which can be attributed to its smaller patch size (4$\times$4). This finer spatial resolution has been shown to improve segmentation quality, albeit at the cost of increased token count and computation. Under the cross-spectral generalization setting, both models exhibit performance degradation, with the complementary-band configuration ($C82$) being the most challenging. We further observe that SpectralViT is especially sensitive to spectral shift ($C120_{\text{SWIR+}}$) and channel expansion ($C202$), where it produces large, spatially coherent misclassifications. In contrast, LESSViT maintains more stable predictions under these shifts. Under simulated EO-1 Hyperion and PRISMA responses, LESSViT better preserves the ground-truth spatial patterns in CDL and BDFORET, demonstrating improved robustness to spectral variation. This difference highlights the limitation of relying on a fixed spectral structure in the encoder, which hinders adaptation under changing spectral distributions, a challenge that LESSViT successfully addresses through flexible spatial–spectral modeling.

\begin{table}[t]
\centering
\small
\setlength{\tabcolsep}{4pt}
\renewcommand{\arraystretch}{1.12}
\caption{\textbf{Comparison of attention operators under spectral configuration shift.} Avg. Seg. denotes the average mIoU across the four segmentation tasks. Bold indicates the best score in each column.}
\label{tab:attention_ablation_main}
\resizebox{0.9\textwidth}{!}{%
\begin{tabular}{lcccccccc}
\toprule
& \multicolumn{4}{c}{Avg. Seg. (mIoU)} & \multicolumn{4}{c}{CORINE (mAP)} \\
\cmidrule(lr){2-5}\cmidrule(lr){6-9}
Attention operator & $C120_{\mathrm{VN+}}$ & $C120_{\mathrm{SW+}}$ & $C82$ & $C202$ & $C120_{\mathrm{VN+}}$ & $C120_{\mathrm{SW+}}$ & $C82$ & $C202$ \\
\midrule
Spatial Attention & \textbf{56.29} & 22.76 & 16.08 & 29.25 & \textbf{76.97} & 29.32 & 26.37 & 33.51 \\
Full Spatial--Spectral Attention & 49.05 & \textbf{46.70} & \textbf{46.49} & \textbf{44.87} & 71.25 & 70.17 & 70.78 & 69.47 \\
\rowcolor[HTML]{ECF4FF}
LESS Attention & 49.19 & 46.07 & 43.97 & 42.88 & 73.19 & \textbf{71.18} & \textbf{72.14} & \textbf{70.00} \\
\bottomrule
\end{tabular}}
\end{table}

\subsection{Effect of Attention Design}
\label{sec:attention_ablation}
To distinguish the effect of attention design from that of pretraining, we compare LESS Attention with full spatial--spectral attention and spatial-only attention under the same HyperMAE pretraining protocol, keeping decoupled masking, hierarchical channel sampling, training data, and optimization settings fixed. All three models use a ViT-S backbone with $64\times64$ inputs to make full spatial--spectral attention computationally feasible.

As shown in \Cref{tab:attention_ablation_main}, spatial-only attention achieves the highest in-distribution scores but degrades substantially under spectral reconfiguration, despite receiving the same spectral exposure during pretraining. Both LESS and full spatial--spectral attention retain substantially stronger performance, supporting the contribution of explicit spatial--spectral modeling to cross-spectral generalization. LESS achieves comparable in-distribution segmentation performance and higher CORINE mAP across all configurations, while full attention performs better on average segmentation mIoU under spectral shifts. These results show that LESS Attention achieves broadly comparable cross-spectral performance to full spatial--spectral attention. We next examine how this performance compares in terms of computational scalability.

\begin{wrapfigure}{r}{0.5\textwidth}
\vspace{-3.5em}
    \centering
    \includegraphics[width=\linewidth]{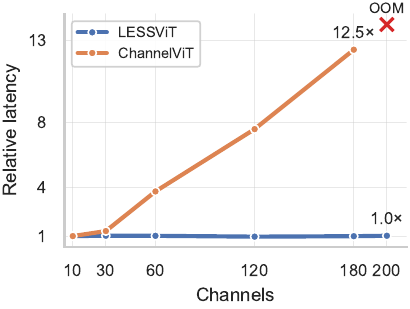}
\caption{
\textbf{Inference latency vs. channel count.}
Latency is normalized to the $C{=}10$ setting. ChannelViT encounters an out-of-memory (OOM) error at $C=200$ on our hardware (144\,GB GPU).
}
\vspace{-1em}
    \label{fig:efficiency}
\end{wrapfigure}
\subsection{Model Efficiency}
We measure the inference latency of LESSViT and ChannelViT~\citep{channel_vit} on 2000 samples while progressively increasing the number of input channels. Latency is normalized to each model's corresponding 10-channel setting, with results shown in \Cref{fig:efficiency}.

As channel count increases, LESSViT maintains near-constant latency, whereas ChannelViT's latency increases rapidly and becomes out of memory at 200 channels under this benchmark configuration. These results demonstrate more favorable scaling with channel count, complementing the broadly comparable cross-spectral performance observed in \Cref{sec:attention_ablation}. Additional measurements of absolute latency, memory usage, and training cost are provided in \Cref{appendix:efficiency}.

\subsection{Impact of Proposed Modules}\label{sec:modules}
We conduct ablations to evaluate key components in LESSViT. All models use a ViT-S-sized backbone and are pretrained with HyperMAE for 50 epochs, with the same training and evaluation protocol as the main experiments. Partial results are shown in \Cref{table:module}, with full results in \Cref{appendix:module}. We study (i) positional encoding, comparing SSRoPE with SIREN~\citep{SIREN}, and (ii) hierarchical channel sampling (HCS) with varying retained ratios $r_\text{HCS}$. 

\begin{table}[t]
\centering
\small
\renewcommand{\arraystretch}{1.05}
\setlength{\tabcolsep}{4pt}
\caption{
Ablation of positional encoding and hierarchical channel sampling (HCS) under cross-spectral generalization. Models are trained on $C120_{\text{VN+}}$ and evaluated across spectral configurations. \textit{Avg. Seg.} denotes the average mIoU across all segmentation tasks.
}
\resizebox{\textwidth}{!}{
\begin{tabular}{ccc cccc cccc cccc}
\toprule
% \multicolumn{2}{c}{Config.}
&&
& \multicolumn{4}{c}{CDL (mIoU)}
& \multicolumn{4}{c}{CORINE (mAP)}
& \multicolumn{4}{c}{\textit{Avg. Seg.} (mIoU)} \\
\cmidrule(lr){4-7} \cmidrule(lr){8-11} \cmidrule(lr){12-15}
Config. & PosEmb & $r_{\text{HCS}}$
& $C120_{\text{VN+}}$ & $C120_{\text{SW+}}$ & $C82$ & $C202$
& $C120_{\text{VN+}}$ & $C120_{\text{SW+}}$ & $C82$ & $C202$
& $C120_{\text{VN+}}$ & $C120_{\text{SW+}}$ & $C82$ & $C202$ \\
\midrule

\bluerow
Default & SSRoPE & $[0.2, 0.3]$ 
&  40.04 &	41.22 &	35.60 & 40.50 
& 70.74	& 68.30 & 	67.11 & 69.61
&  44.48 &	42.66 &	39.96 &	43.63 \\

w/o SSRoPE & SIREN & $[0.2, 0.3]$  
&  28.79 &	29.60 & 30.58 & 29.82
& 70.32	& 69.48	& 64.68	& 69.89
& 40.04	& 37.76 & 37.84 & 38.65 \\

High $r_\text{HCS}$ & SSRoPE  & $[0.4, 0.5]$ 
& 40.93 &	40.87& 35.88 & 39.13
& 69.68	 & 69.04& 64.61	& 69.83
& 45.44	& 44.09 &	40.70 &	43.95 \\

\bottomrule
\end{tabular}
}
\label{table:module}
\end{table}

\textbf{Positional encoding.} Replacing SSRoPE with SIREN degrades performance on segmentation tasks. On in-distribution segmentation tasks, SSRoPE improves spatial prediction quality, suggesting that spatial RoPE provides more effective positional encoding for dense prediction. Under spectral variation, SSRoPE yields more stable performance, indicating that wavelength-aware spectral RoPE supports generalization across varying channel configurations. On CORINE classification, the gap is smaller, suggesting lower sensitivity to positional encoding. Overall, SSRoPE benefits dense spatial modeling in ID settings and supports cross-spectral robustness under OOD channel configurations.

\textbf{Hierarchical channel sampling.}
Increasing the HCS retained ratio from $[0.2, 0.3]$ to $[0.4, 0.5]$ leads to only marginal performance changes across spectral configurations, suggesting that moderate channel sampling does not substantially degrade representation quality. This is likely because HCS primarily reduces the decoder reconstruction space, while the encoder continues to learn from diverse channel subsets through channel masking. At the same time, using the lower $r_\text{HCS}$ reduces pretraining time from 41 hours to 34 hours, corresponding to a 17\% reduction. These results indicate that HCS improves training efficiency with minimal impact on downstream performance.

\section{Related Works}
\textbf{Tensor Product Attention (TPA)}~\citep{tpa}, a concurrent work, introduces a context-aware factorization of attention by constructing query, key, and value tensors through low-rank decompositions. Instead of using fixed projections, TPA dynamically adapts these factorized representations based on the input, enabling expressive attention modeling while maintaining computational efficiency. Although both TPA and our approach lead to similar low-rank formulations of attention, they arise from different motivations and target different objectives. TPA is primarily designed to improve the flexibility and expressiveness of attention through context-aware factorization of query, key, and value representations. In contrast, our method is motivated by the structural separability of spatial and spectral interactions in hyperspectral data, and aims to explicitly model spatial–spectral interactions in a scalable manner. Therefore, while the formulations share similarities, the two approaches are developed under different problem settings and serve distinct purposes.

\section{Conclusions}
In this work, we identify cross-spectral generalization as a fundamental challenge in hyperspectral modeling and present LESSViT as a scalable solution. By combining low-rank spatial–spectral attention, channel-agnostic tokenization, and wavelength-aware positional encoding, LESSViT enables explicit and efficient modeling across varying spectral configurations. A tailored HyperMAE pretraining strategy further makes training tractable under high channel dimensionality. Experiments demonstrate that LESSViT consistently improves robustness under spectral reconfiguration while remaining competitive under in-distribution settings.

Despite these advances, several limitations remain. First, our evaluation uses spectral configurations derived from EnMAP, including simulated PRISMA and Hyperion responses, and therefore does not establish real cross-sensor generalization. Differences in calibration, atmospheric conditions, spatial resolution, and sensor noise remain untested. Second, improving segmentation quality by reducing patch size introduces a substantial increase in token count. Under high token counts and high channel dimensionality, even the linear-complexity formulation of LESS Attention remains computationally demanding, limiting scalability in such regimes.

Future work will focus on validating generalization on real cross-sensor observations and developing more efficient architectures for handling high-resolution spatial–spectral inputs. In addition, our current evaluation is limited to geospatial hyperspectral data. Extending LESSViT to other hyperspectral domains, such as medical imaging and material analysis, and validating its effectiveness on large-scale datasets in these settings remains an important direction.

\begin{ack}
This work was supported by the IBM-Illinois Discovery Accelerator Institute through the IBM-IIDAI joint grant, “Foundation Models for Spatio-temporal and Multi-modal Geospatial Data.” The authors would also like to express their sincere gratitude to Professor Hendrik F. Hamann of Stony Brook University for his valuable guidance and insightful discussions, which helped inspire and shape this work. Computational resources used in this work were partially provided through NSF NAIRR grant NAIRR240419 and the NVIDIA Academic Grant Program.
\end{ack}

\bibliographystyle{plainnat}
\bibliography{reference}

%%%%%%%%%%%%%%%%%%%%%%%%%%%%%%%%%%%%%%%%%%%%%%%%%%%%%%%%%%%%

\newpage
\appendix
\setcounter{table}{0}
\section{Additional Related Works}
\textbf{Hyperspectral Modeling.}
Hyperspectral (HSI) and multispectral (MSI) models for geospatial applications adopt different strategies for capturing spatial--spectral interactions. One line of work aggregates spectral channels during token embedding and subsequently applies attention over spatial tokens. SpectralViT~\citep{spectralearth} projects spectral channels into latent features, while SatMAE~\citep{satmae} uses channel grouping to retain spectral structure. DOFA~\citep{dofa} introduces wavelength-conditioned input parameterization, and HyperFree~\citep{hyperfree} dynamically constructs input embeddings from a learned spectral weight dictionary. Although these designs incorporate spectral information into token embeddings, subsequent attention does not explicitly operate over individual spectral channels.

Another line of work explicitly models spatial--spectral interactions. ChannelViT~\citep{channel_vit} applies attention jointly over spatial and channel tokens, incurring quadratic complexity in the total token count. HyperSigma~\citep{hypersigma} models spatial and spectral dependencies in separate branches and fuses their features. LESSViT preserves explicit joint spatial--spectral modeling while reducing its computational cost through structured low-rank attention.

Complementary work explores spectral pretraining and adaptation to heterogeneous spectral inputs. SpectralMAE~\citep{spectralmae} reconstructs missing spectral bands, while \citet{liu2023self} use masked spectral reconstruction to learn representations for hyperspectral classification. HyperSL~\citep{hypersl} combines masked pretraining with unified spectral tokens and wavelength-aware positional encoding, and HySens~\citep{hysens} harmonizes spectra onto a common wavelength grid with wavelength-based embeddings. These studies motivate learning transferable spectral representations.

\textbf{Hyperspectral Datasets.}
Recent large-scale pretraining datasets and benchmarks such as~\citet{spectralearth,msst,hsihybrid,hypersigma} have significantly expanded the scope of hyperspectral representation learning, introducing data with high spectral dimensionality and varying spatial resolutions. These settings exacerbate the scalability challenges of attention-based models and highlight the need for architectures that can generalize across varying spectral configurations.

\textbf{Tensor Factorization.} 
Low-rank factorization has been widely adopted to reduce computational cost and parameter complexity in deep learning models. A prominent line of work is the LoRA family~\citep{lora}, which decomposes weight updates into low-rank matrices during fine-tuning. Subsequent extensions have explored various training regimes, including efficient pretraining~\citep{relora, jiang2024mora}, long-context modeling~\citep{longlora, sinklora}, and continual learning~\citep{gslora, ilora}. These methods primarily operate on model parameters and construct static low-rank approximations that are independent of the input structure.

\section{Spatial-Spectral RoPE}\label[appendix]{appendix:ssrope}

To encode both spatial information and spectral structure of patches, we introduce spatial–spectral rotary positional embedding (SSRoPE). Our construction is axial: we encode the two spatial coordinates separately and encode the spectral dimension using the physical central wavelength associated with each band. Separate rotary operators are constructed for the spatial and spectral branches and applied within the corresponding branch-specific attention computation.

Let a patch token correspond to spatial position $(u,v)$ and spectral band with central wavelength $\lambda$. For each attention head, we partition the spatial branch into two equal parts corresponding to the horizontal and vertical axes. For frequency index $i$, we define
\begin{equation}
\theta_i^{(u)} = \frac{u}{\tau_i^{(u)}}, \quad 
\theta_i^{(v)} = \frac{v}{\tau_i^{(v)}}, \quad 
\theta_i^{(\lambda)} = \frac{\lambda}{\tau_i^{(\lambda)}},
\end{equation}
where $\tau_i^{(u)}, \tau_i^{(v)}, \tau_i^{(\lambda)}$ are geometric periods.

These phases define block-wise rotations applied to attention features. Denoting by $R(\theta)$ the standard 2D rotation on paired feature dimensions, we construct spatial and spectral rotary operators as
\begin{gather}
R_{\mathrm{s}} = \mathrm{diag}\!\left(
R(\theta_1^{(u)}), \ldots, R(\theta_{d_s/4}^{(u)}),
R(\theta_1^{(v)}), \ldots, R(\theta_{d_s/4}^{(v)})
\right), \\
R_{\mathrm{c}} = \mathrm{diag}\!\left(
R(\theta_1^{(\lambda)}), \ldots, R(\theta_{d_c/2}^{(\lambda)})
\right),
\end{gather}
where $d_s$ and $d_c$ denote the spatial and spectral branch dimensions within each attention head, following the single-head notation in \Cref{sec:less}.

We then apply the rotary transformation to the query and key tensors in each branch:
\begin{align}
\tilde{Q}_s(u,v) = R_{\mathrm{s}}(u,v) Q_s(u,v), &\quad
\tilde{K}_s(u,v) = R_{\mathrm{s}}(u,v) K_s(u,v), \\
\tilde{Q}_c(\lambda) = R_{\mathrm{c}}(\lambda) Q_c(\lambda), &\quad
\tilde{K}_c(\lambda) = R_{\mathrm{c}}(\lambda) K_c(\lambda).
\end{align}

Following standard RoPE~\citep{rope}, the transformation is applied only to non-\texttt{[CLS]} tokens.

\section{Pretraining Details} \label[appendix]{appendix:pretrain}
\subsection{Data}
We pretrain our model using the $C120_{\text{VNIR+}}$ spectral configuration of the SpectralEarth dataset~\citep{spectralearth}, which is a large-scale hyperspectral Earth Observation dataset built from the EnMAP~\citep{enmap} satellite. Each HSI includes 202 channels. The dataset covers 538k locations worldwide. The spatial resolution of the images is 3.84km $\times$ 3.84km. For further details about the dataset, please refer to \citet{spectralearth}.

\subsection{Implementation}
We train our main model for 200 epochs on 4$\times$NVIDIA H200 GPUs, each with 144G VRAM, which takes 96 hours. For comparison, pretraining SpectralViT-B on the same hardware takes 28 hours. LESSViT is trained with an effective batch size of 1024, bfloat16 precision, a base learning rate of 1.5e-4, warmup for 5\% of the total epochs, and cooldown via a cosine annealing scheduler. We optimize the model using the AdamW~\cite{adamw} optimizer with a weight decay of 5e-2. We normalize data following \citet{spectralearth}. For data augmentation, we apply the random horizontal flip with a probability of 50\%. We do not resize the image to preserve the physical property of spatial resolution within the geospatial data.

\section{Evaluation Details}\label[appendix]{appendix:evaluation}
We evaluate our pretrained LESSViT models and other baseline models on released SpectralEarth benchmarks: CDL~\citep{cdl}, EuroCrops~\citep{eurocrops}, CORINE~\citep{corine}, BDFORET~\citep{bdforet} and BNETD~\citep{spectralearth}. Following SpectralEarth's evaluation protocol and our cross-spectral generalization task design, we fine-tune all the models on the $C120_{\text{VNIR+}}$ spectral configuration of the training set, selecting models and hyperparameters based on corresponding validation performance, and reporting the results on the test set for all spectral configurations.

Although downstream training and evaluation follow a common protocol, pretraining data and procedures differ across models. SpectralViT is retrained on $C120_{\mathrm{VNIR+}}$, while DOFA, HyperSigma, and HyperFree use their released checkpoints. These differences remain a confounding factor when attributing cross-model performance differences solely to architecture.

\subsection{Implementation}
As different models and benchmark datasets may have varying optimal learning rates, we conduct a comprehensive search over a wide range of learning rates: \{8e-5, 1e-4, 3e-4, 5e-4, 8e-4\}. This search is applied to all models and benchmark datasets mentioned above, and we report the best results achieved across all learning rates. During fine-tuning, all models use an effective batch size of 256 and bfloat16 precision. The AdamW~\cite{adamw} optimizer is employed for all models with a weight decay of 1e-2, $\beta_{1}=0.9$, and $\beta_{2}=0.999$. We fine-tune all the models for 20 epochs. Additionally, we utilize a cosine annealing scheduler with a warm-up period lasting 20\% of the total fine-tuning epochs to adjust the learning rate during training.

For data augmentation, we start by normalizing all the images. Then we apply random flips both horizontally and vertically, as well as a random rotation. We still avoid any resize operations during data preprocessing, following the data augmentation strategies during pretraining.

% \newpage
\section{Algorithms}
We provide pseudocode for two operations referenced in the main text: full spatial--spectral attention, which serves as a comparison to LESS Attention, and the attention pooling operation used within LESS Attention. The complete LESS Attention block is presented in \Cref{algo:low-rank-spatial-spectral}.

\subsection{Spatial-Spectral Attention Block}\label[appendix]{appendix:ss}
\Cref{algo:spatial-spectral} describes full self-attention over the spatial--spectral token grid, following ChannelViT~\citep{channel_vit}. The $N \times C$ grid is flattened into $NC$ tokens before applying self-attention, resulting in a computational complexity of $\mathcal{O}(N^2C^2D)$.
% \setlength{\textfloatsep}{5pt}
% \begin{algorithm}[t!]
% \caption{Spatial-Spectral Attention Block~\citep{channel_vit}}\label{algo:spatial-spectral}
% \begin{algorithmic}[1]
% \Require Input tokens $X \in \mathbb{R}^{N \times C \times D}$
% \Ensure Output tokens $Y \in \mathbb{R}^{N \times C \times D}$
% \State $\bar{X} \gets \textsc{Reshape}(X, (NC, D)) $ \Comment{Reshape input tokens}
% \State $Q, K, V \gets \bar{X}W_Q, \bar{X}W_K, \bar{X}W_V \in \mathbb{R}^{NC \times D}$ \Comment{Compute query, key, and value}
% \State $A \gets \textsc{Softmax}(QK^\top/\sqrt{d_k})\in \mathbb{R}^{NC \times NC}$ \Comment{Compute attention matrix}
% \State $Y \gets \textsc{Reshape}(AV, (N, C, D))$ \Comment{Apply attention and reshape output}
% \State \Return $Y$
% \end{algorithmic}
% \end{algorithm}

\setlength{\textfloatsep}{5pt}
\begin{algorithm}[h]
    \caption{Spatial-Spectral Attention Block~\cite{channel_vit}}
    \label{algo:spatial-spectral}
    \begin{algorithmic}[1]
        \Require Input tokens $X \in \mathbb{R}^{N \times C \times D}$
        \Ensure Output tokens $Y \in \mathbb{R}^{N \times C \times D}$
        \State $\bar{X} \gets \textsc{Reshape}(X, (NC, D)) $ \Comment{Flatten input tokens}
        \State $Q, K, V \gets \bar{X}W_Q, \bar{X}W_K, \bar{X}W_V \in \mathbb{R}^{NC \times D}$
        \State $Y \gets \textsc{Softmax}(QK^\top/\sqrt{d_k})V\in \mathbb{R}^{NC \times D}$ \Comment{Complexity $O(N^2C^2D)$}
        \State $Y \gets \textsc{Reshape}(Y, (N, C, D))$
        \State \Return $Y$
    \end{algorithmic}
\end{algorithm}

\subsection{Attention Pooling Layer}\label[appendix]{appendix:pma}
\Cref{algo:attentionpool} details the attention pooling operation used to construct the spatial and spectral branches of LESS Attention. Along the axis being pooled, the corresponding \texttt{[CLS]} token supplies the query, while the input tokens supply the keys and values. The resulting weighted aggregation is projected to the target sub-dimension. Applying this operation along the spectral or spatial axis produces the spatial or spectral branch tokens, respectively.
\begin{algorithm}[h]
\caption{AttenPool}
\label{algo:attentionpool}
\begin{algorithmic}[1]
\Require Input tokens $X \in \mathbb{R}^{C \times (N+1) \times D}$, Sub-dimension $d$
\Ensure Pooled tokens $Y \in \mathbb{R}^{C \times 1 \times d}$
  \State \textbf{\# Compute query from the \texttt{[CLS]} tokens of the dimension to be pooled.}
  \State $X_{\texttt{[CLS]}} \leftarrow X[:,\,:1,\,:] \in \mathbb{R}^{C\times 1 \times D}$ \Comment{Extract the attached \texttt{[CLS]} token.}
  \State $Q \leftarrow X_{\texttt{[CLS]}}W_Q \in \mathbb{R}^{C\times 1 \times D} $
  \State \textbf{\# Compute key and value from the input tokens.}
  \State $K, V \leftarrow XW_K,XW_V \in \mathbb{R}^{C\times (N+1) \times D} $
  \State \textbf{\# Use a weighted sum to pool the targeting dimension.}
  \State $A \gets \textsc{Softmax}(QK^\top/\sqrt{d_k}) \in \mathbb{R}^{C \times 1 \times (N+1)}$ \Comment{The weight is essentially the attention map.}
  \State $\bar Y \gets AV \in \mathbb{R}^{C \times 1 \times D}$
  \State $Y \gets \bar YW_d \in \mathbb{R}^{C \times 1 \times d}$ \Comment{Project to the desired sub-dimension.}
  \State \Return $Y$
\end{algorithmic}
\end{algorithm}

\section{Additional Experiment Results}

\subsection{Full Fine-tune and Linear-probing} \label[appendix]{appendix:lpft}
We report full fine-tuning (FT) and linear probing (LP) results across five downstream datasets and six spectral configurations in \Cref{tab:full_ft_lp}. FT updates the encoder and task head, whereas LP freezes the pretrained encoder and trains only a linear prediction head. Both protocols use $C120_{\mathrm{VNIR+}}$ for downstream training and evaluate the resulting models across all configurations without further adaptation. Results are reported as mean $\pm$ standard deviation across five random seeds.

Under FT, LESSViT achieves the highest mean performance across all 25 out-of-distribution task--configuration pairs. The multi-seed results support the cross-spectral robustness observed in the main experiments, while SpectralViT retains the strongest in-distribution performance. LP further highlights this specialization--robustness trade-off: SpectralViT performs best in distribution but degrades sharply under spectral shifts, whereas LESSViT achieves the highest mean score in 17 of the 25 out-of-distribution pairs. Since LP freezes the encoder, these results indicate that LESSViT's robustness is already encoded in its pretrained representations, rather than arising solely from downstream fine-tuning.
% Complete replacement for tables/full_results.tex.
% Requires longtable, array, booktabs, amsmath, and xcolor with table support.
% Keeps all original means, standard deviations, drops, and bold markings.
\begingroup
\scriptsize
\renewcommand{\arraystretch}{1.1}
\setlength{\tabcolsep}{2pt}
\setlength{\LTcapwidth}{\textwidth}
\setlength{\LTleft}{0pt}
\setlength{\LTright}{0pt}
% First column + six result columns + twelve internal gaps = \textwidth.
\newcommand{\FullResultCell}[3]{%
  \shortstack[c]{$#1\pm #2$\\[-1pt]$\scriptstyle\downarrow #3\%$}}
\begin{longtable}{@{}p{0.13\textwidth}*{6}{>{\centering\arraybackslash}p{\dimexpr(\textwidth-0.13\textwidth-12\tabcolsep)/6\relax}}@{}}
\caption{\textbf{Full fine-tuning (FT) and linear probing (LP) results across five downstream tasks.} Scores are means $\pm$ standard deviations over five seeds; percentages indicate relative drops from the corresponding in-distribution configuration. Bold denotes the highest mean within each dataset--configuration pair under each evaluation protocol.}
\label{tab:full_ft_lp}\\
\toprule
Model & $C120_{\mathrm{VN+}}$ & $C120_{\mathrm{SW+}}$ & $C82$ & $C202$ & PRISMA & EO1H \\
\midrule
\endfirsthead
\multicolumn{7}{c}{\tablename~\thetable{} (continued)} \\
\toprule
Model & $C120_{\mathrm{VN+}}$ & $C120_{\mathrm{SW+}}$ & $C82$ & $C202$ & PRISMA & EO1H \\
\midrule
\endhead
\midrule
\multicolumn{7}{r}{\textit{Continued on next page}} \\
\endfoot
\bottomrule
\endlastfoot
\multicolumn{7}{c}{\textit{CDL (mIoU) --- Full fine-tuning (FT)}} \\*
\midrule
SpectralViT & $\mathbf{73.33}\pm 1.70$ & \FullResultCell{15.11}{1.13}{79.4} & \FullResultCell{12.36}{3.15}{83.1} & \FullResultCell{5.59}{0.17}{92.4} & \FullResultCell{10.84}{2.51}{85.2} & \FullResultCell{7.81}{1.03}{89.4} \\*
HyperSigma & $44.48\pm 3.62$ & \FullResultCell{13.93}{1.24}{68.7} & \FullResultCell{14.50}{2.91}{67.4} & \FullResultCell{14.51}{0.54}{67.4} & \FullResultCell{10.08}{3.01}{77.3} & \FullResultCell{12.86}{3.02}{71.1} \\*
DOFA & $40.06\pm 1.85$ & \FullResultCell{18.38}{0.18}{54.1} & \FullResultCell{18.08}{0.75}{54.9} & \FullResultCell{12.39}{0.81}{69.1} & \FullResultCell{13.06}{1.54}{67.4} & \FullResultCell{6.39}{4.10}{84.1} \\*
HyperFree & $42.50\pm 0.62$ & \FullResultCell{28.48}{1.95}{33.0} & \FullResultCell{16.22}{2.44}{61.8} & \FullResultCell{22.16}{0.70}{47.9} & \FullResultCell{28.09}{1.71}{33.9} & \FullResultCell{19.59}{3.47}{53.9} \\*
\rowcolor[HTML]{ECF4FF}
LESSViT & $53.44\pm 1.02$ & \FullResultCell{\mathbf{48.32}}{5.26}{9.6} & \FullResultCell{\mathbf{42.08}}{5.21}{21.3} & \FullResultCell{\mathbf{50.75}}{2.65}{5.0} & \FullResultCell{\mathbf{48.96}}{3.42}{8.4} & \FullResultCell{\mathbf{49.05}}{4.53}{8.2} \\
\midrule
\multicolumn{7}{c}{\textit{CDL (mIoU) --- Linear probing (LP)}} \\*
\midrule
SpectralViT & $\mathbf{60.65}\pm 0.78$ & \FullResultCell{3.75}{0.40}{93.8} & \FullResultCell{3.14}{0.29}{94.8} & \FullResultCell{6.44}{0.23}{89.4} & \FullResultCell{3.71}{0.33}{93.9} & \FullResultCell{3.77}{0.25}{93.8} \\*
HyperSigma & $27.32\pm 2.21$ & \FullResultCell{18.18}{1.71}{33.5} & \FullResultCell{2.26}{0.51}{91.7} & \FullResultCell{2.09}{0.08}{92.3} & \FullResultCell{6.81}{0.17}{75.1} & \FullResultCell{12.40}{0.82}{54.6} \\*
DOFA & $31.81\pm 1.44$ & \FullResultCell{4.59}{0.04}{85.6} & \FullResultCell{4.08}{0.20}{87.2} & \FullResultCell{4.14}{0.01}{87.0} & \FullResultCell{6.24}{0.95}{80.4} & \FullResultCell{3.67}{0.53}{88.5} \\*
HyperFree & $30.12\pm 0.62$ & \FullResultCell{18.99}{0.68}{36.9} & \FullResultCell{15.14}{2.79}{49.7} & \FullResultCell{7.06}{0.23}{76.6} & \FullResultCell{21.46}{0.18}{28.8} & \FullResultCell{22.62}{0.58}{24.9} \\*
\rowcolor[HTML]{ECF4FF}
LESSViT & $46.29\pm 0.51$ & \FullResultCell{\mathbf{31.43}}{3.93}{32.1} & \FullResultCell{\mathbf{30.29}}{3.30}{34.6} & \FullResultCell{\mathbf{39.34}}{1.32}{15.0} & \FullResultCell{\mathbf{29.11}}{0.20}{37.1} & \FullResultCell{\mathbf{31.01}}{0.23}{33.0} \\
\midrule
\multicolumn{7}{c}{\textit{BDFORET (mIoU) --- Full fine-tuning (FT)}} \\*
\midrule
SpectralViT & $\mathbf{77.17}\pm 0.50$ & \FullResultCell{48.31}{0.46}{37.4} & \FullResultCell{48.68}{0.45}{36.9} & \FullResultCell{48.73}{0.21}{36.9} & \FullResultCell{46.77}{1.04}{39.4} & \FullResultCell{44.55}{2.40}{42.3} \\*
HyperSigma & $63.76\pm 0.33$ & \FullResultCell{47.97}{1.32}{24.8} & \FullResultCell{44.15}{3.64}{30.8} & \FullResultCell{48.94}{1.13}{23.2} & \FullResultCell{43.68}{2.62}{31.5} & \FullResultCell{9.54}{2.19}{85.0} \\*
DOFA & $58.14\pm 0.50$ & \FullResultCell{51.26}{0.63}{11.8} & \FullResultCell{42.03}{4.75}{27.7} & \FullResultCell{50.45}{0.19}{13.2} & \FullResultCell{45.04}{3.44}{22.5} & \FullResultCell{2.36}{1.37}{95.9} \\*
HyperFree & $58.42\pm 2.09$ & \FullResultCell{54.43}{2.85}{6.8} & \FullResultCell{52.63}{3.02}{9.9} & \FullResultCell{54.13}{1.59}{7.3} & \FullResultCell{54.44}{1.55}{6.8} & \FullResultCell{53.93}{1.60}{7.7} \\*
\rowcolor[HTML]{ECF4FF}
LESSViT & $63.87\pm 0.61$ & \FullResultCell{\mathbf{62.43}}{2.02}{2.3} & \FullResultCell{\mathbf{60.58}}{3.68}{5.2} & \FullResultCell{\mathbf{63.35}}{0.54}{0.8} & \FullResultCell{\mathbf{62.03}}{0.83}{2.9} & \FullResultCell{\mathbf{62.60}}{0.55}{2.0} \\
\midrule
\multicolumn{7}{c}{\textit{BDFORET (mIoU) --- Linear probing (LP)}} \\*
\midrule
SpectralViT & $\mathbf{66.58}\pm 0.25$ & \FullResultCell{46.70}{0.08}{29.9} & \FullResultCell{46.23}{0.10}{30.6} & \FullResultCell{33.75}{0.31}{49.3} & \FullResultCell{47.37}{0.06}{28.9} & \FullResultCell{41.60}{1.90}{37.5} \\*
HyperSigma & $50.75\pm 0.44$ & \FullResultCell{45.58}{3.03}{10.2} & \FullResultCell{6.67}{0.00}{86.9} & \FullResultCell{46.71}{3.25}{8.0} & \FullResultCell{21.54}{2.90}{57.6} & \FullResultCell{48.80}{0.00}{3.8} \\*
DOFA & $55.75\pm 0.76$ & \FullResultCell{4.26}{0.03}{92.4} & \FullResultCell{4.19}{0.08}{92.5} & \FullResultCell{3.88}{0.51}{93.0} & \FullResultCell{3.62}{0.03}{93.5} & \FullResultCell{3.55}{0.06}{93.6} \\*
HyperFree & $53.84\pm 0.30$ & \FullResultCell{\mathbf{49.08}}{1.34}{8.8} & \FullResultCell{50.11}{0.68}{6.9} & \FullResultCell{49.86}{1.56}{7.4} & \FullResultCell{\mathbf{50.10}}{0.05}{6.9} & \FullResultCell{\mathbf{49.10}}{0.08}{8.8} \\*
\rowcolor[HTML]{ECF4FF}
LESSViT & $59.60\pm 0.02$ & \FullResultCell{42.31}{0.71}{29.0} & \FullResultCell{\mathbf{51.35}}{2.56}{13.9} & \FullResultCell{\mathbf{53.26}}{0.23}{10.6} & \FullResultCell{45.85}{0.33}{23.1} & \FullResultCell{45.76}{0.36}{23.2} \\
\midrule
\multicolumn{7}{c}{\textit{BNETD (mIoU) --- Full fine-tuning (FT)}} \\*
\midrule
SpectralViT & $\mathbf{44.69}\pm 0.64$ & \FullResultCell{16.86}{2.59}{62.3} & \FullResultCell{17.06}{1.01}{61.8} & \FullResultCell{16.66}{1.33}{62.7} & \FullResultCell{18.00}{1.42}{59.7} & \FullResultCell{16.07}{3.03}{64.1} \\*
HyperSigma & $30.36\pm 2.96$ & \FullResultCell{18.78}{0.90}{38.1} & \FullResultCell{17.64}{0.73}{41.9} & \FullResultCell{17.68}{1.98}{41.8} & \FullResultCell{15.31}{0.17}{49.6} & \FullResultCell{7.22}{0.38}{76.2} \\*
DOFA & $28.88\pm 2.20$ & \FullResultCell{18.06}{0.04}{37.5} & \FullResultCell{18.87}{0.30}{34.7} & \FullResultCell{18.89}{0.23}{34.6} & \FullResultCell{15.75}{0.18}{45.5} & \FullResultCell{6.29}{0.48}{78.2} \\*
HyperFree & $29.53\pm 0.86$ & \FullResultCell{18.53}{2.19}{37.3} & \FullResultCell{16.87}{1.93}{42.9} & \FullResultCell{21.47}{1.98}{27.3} & \FullResultCell{21.93}{0.53}{25.7} & \FullResultCell{18.97}{0.28}{35.8} \\*
\rowcolor[HTML]{ECF4FF}
LESSViT & $32.37\pm 0.82$ & \FullResultCell{\mathbf{30.76}}{2.57}{5.0} & \FullResultCell{\mathbf{26.35}}{4.28}{18.6} & \FullResultCell{\mathbf{31.25}}{1.85}{3.5} & \FullResultCell{\mathbf{31.80}}{0.30}{1.8} & \FullResultCell{\mathbf{29.62}}{3.07}{8.5} \\
\midrule
\multicolumn{7}{c}{\textit{BNETD (mIoU) --- Linear probing (LP)}} \\*
\midrule
SpectralViT & $\mathbf{37.55}\pm 0.63$ & \FullResultCell{10.40}{0.39}{72.3} & \FullResultCell{11.38}{0.39}{69.7} & \FullResultCell{10.67}{0.51}{71.6} & \FullResultCell{10.41}{0.08}{72.3} & \FullResultCell{9.55}{0.04}{74.6} \\*
HyperSigma & $21.35\pm 0.71$ & \FullResultCell{12.58}{0.50}{41.1} & \FullResultCell{5.41}{0.01}{74.7} & \FullResultCell{8.62}{5.36}{59.6} & \FullResultCell{5.22}{1.05}{75.5} & \FullResultCell{8.04}{0.42}{62.4} \\*
DOFA & $24.54\pm 0.40$ & \FullResultCell{8.44}{1.31}{65.6} & \FullResultCell{5.10}{0.49}{79.2} & \FullResultCell{14.80}{0.68}{39.7} & \FullResultCell{8.27}{0.81}{66.3} & \FullResultCell{8.10}{1.16}{67.0} \\*
HyperFree & $24.84\pm 0.54$ & \FullResultCell{15.64}{0.65}{37.0} & \FullResultCell{\mathbf{16.76}}{2.33}{32.5} & \FullResultCell{12.71}{0.03}{48.8} & \FullResultCell{11.75}{0.67}{52.7} & \FullResultCell{\mathbf{19.59}}{0.32}{21.2} \\*
\rowcolor[HTML]{ECF4FF}
LESSViT & $30.11\pm 0.08$ & \FullResultCell{\mathbf{17.73}}{0.49}{41.1} & \FullResultCell{14.42}{0.40}{52.1} & \FullResultCell{\mathbf{23.05}}{0.52}{23.4} & \FullResultCell{\mathbf{17.00}}{1.07}{43.5} & \FullResultCell{18.20}{1.40}{39.6} \\
\midrule
\multicolumn{7}{c}{\textit{EuroCrops (mIoU) --- Full fine-tuning (FT)}} \\*
\midrule
SpectralViT & $\mathbf{71.69}\pm 0.79$ & \FullResultCell{26.27}{5.33}{63.4} & \FullResultCell{41.14}{1.13}{42.6} & \FullResultCell{46.28}{1.07}{35.4} & \FullResultCell{42.23}{2.19}{41.1} & \FullResultCell{29.32}{10.24}{59.1} \\*
HyperSigma & $56.54\pm 0.28$ & \FullResultCell{52.16}{1.17}{7.7} & \FullResultCell{48.45}{3.36}{14.3} & \FullResultCell{51.59}{0.14}{8.8} & \FullResultCell{46.32}{2.57}{18.1} & \FullResultCell{38.84}{2.96}{31.3} \\*
DOFA & $55.59\pm 0.63$ & \FullResultCell{36.85}{5.04}{33.7} & \FullResultCell{32.83}{1.94}{40.9} & \FullResultCell{31.52}{3.17}{43.3} & \FullResultCell{50.28}{0.27}{9.5} & \FullResultCell{50.40}{0.004}{9.3} \\*
HyperFree & $52.22\pm 2.10$ & \FullResultCell{46.63}{1.09}{10.7} & \FullResultCell{32.87}{6.30}{37.1} & \FullResultCell{36.25}{7.10}{30.6} & \FullResultCell{45.71}{2.00}{12.5} & \FullResultCell{20.91}{6.14}{60.0} \\*
\rowcolor[HTML]{ECF4FF}
LESSViT & $57.91\pm 0.61$ & \FullResultCell{\mathbf{55.69}}{5.11}{3.8} & \FullResultCell{\mathbf{55.47}}{2.48}{4.2} & \FullResultCell{\mathbf{57.81}}{1.64}{0.2} & \FullResultCell{\mathbf{57.63}}{1.42}{0.5} & \FullResultCell{\mathbf{57.48}}{1.17}{0.7} \\
\midrule
\multicolumn{7}{c}{\textit{EuroCrops (mIoU) --- Linear probing (LP)}} \\*
\midrule
SpectralViT & $\mathbf{65.62}\pm 0.07$ & \FullResultCell{28.27}{0.10}{56.9} & \FullResultCell{12.30}{0.11}{81.3} & \FullResultCell{31.87}{0.08}{51.4} & \FullResultCell{31.27}{1.69}{52.3} & \FullResultCell{\mathbf{46.43}}{0.33}{29.2} \\*
HyperSigma & $51.13\pm 0.01$ & \FullResultCell{\mathbf{50.43}}{0.28}{1.4} & \FullResultCell{23.06}{2.71}{54.9} & \FullResultCell{27.53}{0.20}{46.2} & \FullResultCell{34.08}{1.14}{33.3} & \FullResultCell{43.16}{0.45}{15.6} \\*
DOFA & $53.70\pm 0.34$ & \FullResultCell{30.11}{5.26}{43.9} & \FullResultCell{16.80}{3.03}{68.7} & \FullResultCell{38.76}{2.42}{27.8} & \FullResultCell{\mathbf{47.08}}{2.98}{12.3} & \FullResultCell{43.77}{3.44}{18.5} \\*
HyperFree & $51.04\pm 0.03$ & \FullResultCell{33.68}{3.69}{34.0} & \FullResultCell{29.06}{3.82}{43.1} & \FullResultCell{17.76}{3.21}{65.2} & \FullResultCell{43.69}{3.66}{14.4} & \FullResultCell{35.89}{4.66}{29.7} \\*
\rowcolor[HTML]{ECF4FF}
LESSViT & $54.58\pm 3.73$ & \FullResultCell{45.29}{4.55}{17.0} & \FullResultCell{\mathbf{44.11}}{5.32}{19.2} & \FullResultCell{\mathbf{51.44}}{1.81}{5.8} & \FullResultCell{45.23}{1.42}{17.1} & \FullResultCell{45.86}{3.50}{16.0} \\
\midrule
\multicolumn{7}{c}{\textit{CORINE (mAP) --- Full fine-tuning (FT)}} \\*
\midrule
SpectralViT & $\mathbf{78.93}\pm 0.19$ & \FullResultCell{36.90}{1.28}{53.2} & \FullResultCell{36.92}{2.15}{53.2} & \FullResultCell{38.37}{1.07}{51.4} & \FullResultCell{35.24}{2.59}{55.4} & \FullResultCell{33.08}{1.81}{58.1} \\*
HyperSigma & $69.21\pm 0.52$ & \FullResultCell{36.05}{4.65}{47.9} & \FullResultCell{27.56}{1.06}{60.2} & \FullResultCell{47.11}{2.57}{31.9} & \FullResultCell{38.60}{4.03}{44.2} & \FullResultCell{39.48}{0.91}{43.0} \\*
DOFA & $66.40\pm 6.96$ & \FullResultCell{37.89}{2.99}{42.9} & \FullResultCell{48.27}{2.94}{27.3} & \FullResultCell{35.33}{1.15}{46.8} & \FullResultCell{25.66}{7.74}{61.4} & \FullResultCell{17.58}{0.19}{73.5} \\*
HyperFree & $74.04\pm 0.76$ & \FullResultCell{47.93}{2.67}{35.3} & \FullResultCell{63.31}{4.67}{14.5} & \FullResultCell{68.48}{0.22}{7.5} & \FullResultCell{65.83}{0.59}{11.1} & \FullResultCell{62.28}{1.50}{15.9} \\*
\rowcolor[HTML]{ECF4FF}
LESSViT & $76.35\pm 0.46$ & \FullResultCell{\mathbf{74.21}}{3.55}{2.8} & \FullResultCell{\mathbf{72.95}}{4.05}{4.5} & \FullResultCell{\mathbf{75.99}}{0.30}{0.5} & \FullResultCell{\mathbf{75.05}}{0.82}{1.7} & \FullResultCell{\mathbf{75.01}}{1.43}{1.8} \\
\midrule
\multicolumn{7}{c}{\textit{CORINE (mAP) --- Linear probing (LP)}} \\*
\midrule
SpectralViT & $\mathbf{71.33}\pm 0.21$ & \FullResultCell{37.55}{0.06}{47.4} & \FullResultCell{39.40}{0.07}{44.8} & \FullResultCell{37.05}{0.47}{48.1} & \FullResultCell{49.42}{0.10}{30.7} & \FullResultCell{34.64}{0.08}{51.4} \\*
HyperSigma & $61.66\pm 0.14$ & \FullResultCell{41.76}{1.65}{32.3} & \FullResultCell{24.15}{1.10}{60.8} & \FullResultCell{35.01}{0.39}{43.2} & \FullResultCell{42.79}{0.04}{30.6} & \FullResultCell{40.93}{1.91}{33.6} \\*
DOFA & $52.18\pm 1.16$ & \FullResultCell{19.72}{0.61}{62.2} & \FullResultCell{11.05}{0.52}{78.8} & \FullResultCell{30.05}{0.71}{42.4} & \FullResultCell{30.88}{0.68}{40.8} & \FullResultCell{29.64}{0.17}{43.2} \\*
HyperFree & $63.11\pm 0.08$ & \FullResultCell{49.32}{0.03}{21.8} & \FullResultCell{45.24}{1.14}{28.3} & \FullResultCell{43.61}{0.50}{30.9} & \FullResultCell{46.10}{1.07}{27.0} & \FullResultCell{50.31}{0.08}{20.3} \\*
\rowcolor[HTML]{ECF4FF}
LESSViT & $69.96\pm 0.02$ & \FullResultCell{\mathbf{60.26}}{1.07}{13.9} & \FullResultCell{\mathbf{57.67}}{0.91}{17.6} & \FullResultCell{\mathbf{63.97}}{0.96}{8.6} & \FullResultCell{\mathbf{61.67}}{0.69}{11.9} & \FullResultCell{\mathbf{60.56}}{0.88}{13.4} \\
\end{longtable}
\endgroup

\subsection{Effective Rank of Pretrained Representations} \label[appendix]{appendix:feature_rank}
To examine whether the low-rank structure of LESS Attention limits the dimensionality of learned representations, we measure the effective rank of pretrained encoder features using RankMe~\citep{rankme}, normalized by the embedding dimension ($D=768$). HyperSigma is excluded because its dual branches are fused only at the task head and do not provide a unified intermediate representation for this comparison.

As shown in \Cref{tab:effective_rank}, LESSViT exhibits the highest normalized effective rank across all four spectral configurations. These results indicate that its structured low-rank attention does not lead to low-rank encoder features. Instead, LESSViT learns representations that occupy a larger effective fraction of the embedding space than the compared models.
\begin{table}[h]
\centering
\small
\setlength{\tabcolsep}{7pt}
\setlength{\belowcaptionskip}{6pt}
\renewcommand{\arraystretch}{1.15}
\caption{\textbf{Effective rank across spectral configurations.} Bold indicates the highest value in each configuration.}
\label{tab:effective_rank}
\resizebox{0.5\textwidth}{!}{%
\begin{tabular}{lcccc}
\toprule
Model & $C120_{\mathrm{VNIR+}}$ & $C120_{\mathrm{SWIR+}}$ & $C82$ & $C202$ \\
\midrule
SpectralViT & 0.255 & 0.275 & 0.259 & 0.275 \\
DOFA & 0.092 & 0.087 & 0.083 & 0.091 \\
HyperFree & 0.260 & 0.244 & 0.246 & 0.245 \\
\rowcolor[HTML]{ECF4FF}
LESSViT & \textbf{0.343} & \textbf{0.362} & \textbf{0.361} & \textbf{0.358} \\
\bottomrule
\end{tabular}%
}
\end{table}

\subsection{In-Distribution Performance and Cross-Spectral Robustness} \label[appendix]{appendix:id_ood}
To quantify the trade-off between in-distribution specialization and cross-spectral robustness, we report ID performance, average OOD performance, and retention rates in \Cref{tab:id_ood_tradeoff}, computed from multi-seed fine-tuning means. ID denotes $C120_{\mathrm{VNIR+}}$, while OOD averages over $C120_{\mathrm{SWIR+}}$, $C82$, $C202$, and the simulated PRISMA and Hyperion configurations. Segmentation mIoU is averaged across CDL, BDFORET, BNETD, and EuroCrops; classification mAP is reported on CORINE. Retention is the ratio of aggregated OOD to ID performance, expressed as a percentage.

SpectralViT achieves the strongest in-distribution performance but retains only 41.86\% of its segmentation mIoU and 45.74\% of its classification mAP under spectral shifts. In contrast, LESSViT retains 94.80\% and 97.76\%, respectively, while achieving the highest average OOD performance for both metrics. These results highlight the trade-off between fixed-configuration specialization and cross-spectral robustness, with LESSViT providing a more favorable balance.
\begin{table}[h]
\centering
\small
\setlength{\tabcolsep}{5pt}
\renewcommand{\arraystretch}{1.15}
\caption{\textbf{In-distribution performance and out-of-distribution robustness.} Bold indicates the best result in each column.}
\label{tab:id_ood_tradeoff}
\resizebox{0.70\textwidth}{!}{%
\begin{tabular}{lcccccc}
\toprule
& \multicolumn{3}{c}{Segmentation (mIoU)} & \multicolumn{3}{c}{Classification (mAP)} \\
\cmidrule(lr){2-4}\cmidrule(lr){5-7}
Model & ID & Avg. OOD & Retention (\%) & ID & Avg. OOD & Retention (\%) \\
\midrule
SpectralViT & \textbf{66.72} & 27.93 & 41.86 & \textbf{78.93} & 36.10 & 45.74 \\
HyperSigma & 48.78 & 28.71 & 58.84 & 69.21 & 37.76 & 54.56 \\
DOFA & 45.67 & 26.96 & 59.03 & 66.40 & 32.95 & 49.62 \\
HyperFree & 45.67 & 33.21 & 72.73 & 74.04 & 61.57 & 83.15 \\
\rowcolor[HTML]{ECF4FF}
LESSViT & 51.90 & \textbf{49.20} & \textbf{94.80} & 76.35 & \textbf{74.64} & \textbf{97.76} \\
\bottomrule
\end{tabular}%
}
\end{table}

\subsection{Attention and Component Ablations}
\label[appendix]{appendix:ablations}

\paragraph{Attention operators.}
\label[appendix]{appendix:attention_abl}
We provide the per-dataset results of the controlled attention comparison in \Cref{tab:attention_ablation}. All three models use a ViT-S backbone with $64\times64$ inputs and identical HyperMAE pretraining, including decoupled masking, hierarchical channel sampling, training data, and optimization settings. Spatial-only attention achieves strong in-distribution performance but degrades sharply under spectral reconfiguration, showing that broader spectral exposure during pretraining alone does not explain the robustness gains. LESS and full spatial--spectral attention achieve broadly comparable performance, with neither consistently outperforming the other, supporting LESS as a scalable alternative to full attention.
\begin{table}[t]
\centering
\small
\setlength{\tabcolsep}{4pt}
\renewcommand{\arraystretch}{1.12}
\caption{\textbf{Comparison of attention operators under spectral configuration shift.} Results are reported as mIoU for segmentation and mAP for CORINE classification. Bold indicates the best score in each column.}
\label{tab:attention_ablation}
\resizebox{1\textwidth}{!}{%
\begin{tabular}{lcccccccccccc}
\toprule
 & \multicolumn{4}{c}{CDL (mIoU)} & \multicolumn{4}{c}{EuroCrops (mIoU)} & \multicolumn{4}{c}{CORINE (mAP)} \\
\cmidrule(lr){2-5}\cmidrule(lr){6-9}\cmidrule(lr){10-13}
Attention operator & $C120_{\mathrm{VN+}}$ & $C120_{\mathrm{SW+}}$ & $C82$ & $C202$ & $C120_{\mathrm{VN+}}$ & $C120_{\mathrm{SW+}}$ & $C82$ & $C202$ & $C120_{\mathrm{VN+}}$ & $C120_{\mathrm{SW+}}$ & $C82$ & $C202$ \\
\midrule
Spatial Attention & \textbf{58.92} & 7.41 & 3.20 & 15.52 & \textbf{62.12} & 42.11 & 30.79 & 49.71 & \textbf{76.97} & 29.32 & 26.37 & 33.51 \\
Full Spatial--Spectral Attention & 45.04 & 41.95 & 42.64 & 37.66 & 57.84 & \textbf{57.40} & \textbf{57.34} & \textbf{57.44} & 71.25 & 70.17 & 70.78 & 69.47 \\
\rowcolor[HTML]{ECF4FF}
LESS Attention & 44.22 & \textbf{42.44} & \textbf{42.85} & \textbf{40.47} & 57.72 & 56.15 & 56.10 & 55.73 & 73.19 & \textbf{71.18} & \textbf{72.14} & \textbf{70.00} \\
\bottomrule
\end{tabular}}
\par\vspace{6pt}
\resizebox{0.76\textwidth}{!}{%
\begin{tabular}{lcccccccc}
\toprule
 & \multicolumn{4}{c}{BDFORET (mIoU)} & \multicolumn{4}{c}{BNETD (mIoU)} \\
\cmidrule(lr){2-5}\cmidrule(lr){6-9}
Attention operator & $C120_{\mathrm{VN+}}$ & $C120_{\mathrm{SW+}}$ & $C82$ & $C202$ & $C120_{\mathrm{VN+}}$ & $C120_{\mathrm{SW+}}$ & $C82$ & $C202$ \\
\midrule
Spatial Attention & \textbf{68.26} & 32.12 & 21.87 & 48.10 & \textbf{35.87} & 9.39 & 8.46 & 3.68 \\
Full Spatial--Spectral Attention & 62.81 & \textbf{58.85} & 58.53 & \textbf{58.25} & 30.51 & \textbf{28.61} & \textbf{27.45} & \textbf{26.11} \\
\rowcolor[HTML]{ECF4FF}
LESS Attention & 63.51 & 57.75 & \textbf{59.92} & 56.60 & 31.29 & 27.95 & 17.01 & 18.72 \\
\bottomrule
\end{tabular}}
\end{table}

\paragraph{Positional encoding and channel sampling.}
\label[appendix]{appendix:module}
\Cref{tab:module_full} reports the full component-ablation results. SSRoPE improves average segmentation performance over SIREN across all spectral configurations. Increasing the HCS retained ratio $r_\text{HCS}$ from $[0.2,0.3]$ to $[0.4,0.5]$ yields modest changes in downstream performance, while the lower ratio reduces pretraining time from 41 to 34 hours. Tied patch embedding and decoupled masking are structural requirements of the proposed formulation: the former supports varying channel counts, while the latter preserves the axis-aligned token grid required by the factorized attention.
\begin{table}[h]
\centering
\small
\renewcommand{\arraystretch}{1.05}
\setlength{\tabcolsep}{3.2pt}

\caption{
\textbf{Ablation of key modules under cross-spectral generalization.} Models are trained on $C120_{\text{VN+}}$ and evaluated across spectral configurations ($C120_{\text{VN+}}$, $C120_{\text{SW+}}$, $C82$, $C202$). We compare positional encoding (SSRoPE vs. SIREN) and hierarchical channel sampling (HCS) ratios. Metrics include mIoU for segmentation tasks and mAP for classification tasks.
}
\label{tab:module_full}

% ================= Top Block =================
\begin{minipage}{\textwidth}
\centering
\resizebox{\textwidth}{!}{
\begin{tabular}{ccc cccc cccc cccc}
\toprule
% \multicolumn{2}{c}{Config.}
&&
& \multicolumn{4}{c}{CDL (mIoU)}
& \multicolumn{4}{c}{EuroCrops (mIoU)}
& \multicolumn{4}{c}{CORINE (mAP)}\\
\cmidrule(lr){4-7} \cmidrule(lr){8-11} \cmidrule(lr){12-15}
Config. & PE & $r_{\text{HCS}}$
& $C120_{\text{VN+}}$ & $C120_{\text{SW+}}$ & $C82$ & $C202$
& $C120_{\text{VN+}}$ & $C120_{\text{SW+}}$ & $C82$ & $C202$
& $C120_{\text{VN+}}$ & $C120_{\text{SW+}}$ & $C82$ & $C202$ \\
\midrule

\bluerow
Default & SSRoPE & $[0.2, 0.3]$ 
&  40.04 &	41.22 &	35.60 & 40.50 
& 52.52	& 46.94	& 47.49	& 49.78
& 70.74	& 68.30 & 	67.11 & 69.61\\

w/o SSRoPE & SIREN & $[0.2, 0.3]$  
& 28.79	& 29.60	&30.58&	29.82
& 50.45	& 45.77	& 46.01	& 47.27 
& 70.32	& 69.48	& 64.68	& 69.89\\

High $r_\text{HCS}$ & SSRoPE  & $[0.4, 0.5]$ 
& 40.93	& 40.87	& 35.88	& 39.13
& 52.06	& 51.54	& 48.70	& 51.37
& 69.68	& 69.04	& 64.61	& 69.83 \\

\bottomrule
\end{tabular}
}
\end{minipage}
\vspace{6pt}

% ================= Bottom Block =================
\begin{minipage}{\textwidth}
\centering
\resizebox{0.5\textwidth}{!}{
\begin{tabular}{cccc cccc}
\toprule
\multicolumn{4}{c}{BDFORET (mIoU)}
& \multicolumn{4}{c}{BNETD (mIoU)} \\
\cmidrule(lr){1-4}\cmidrule(lr){5-8}

$C120_{\text{VN+}}$ & $C120_{\text{SW+}}$ & $C82$ & $C202$
& $C120_{\text{VN+}}$ & $C120_{\text{SW+}}$ & $C82$ & $C202$ \\
\midrule

\bluerow
57.30	& 58.22	& 54.39	& 58.11
& 28.04	& 24.24	& 22.37	& 26.14\\

55.51& 54.14& 53.89& 54.41
& 25.41	& 21.51	 & 20.89& 23.08\\

59.81&	58.19	&54.45	&58.85
&28.95&	25.77	&23.76	&26.44\\

\bottomrule
\end{tabular}
}
\end{minipage}

\end{table}

\subsection{Computational Efficiency}
\label[appendix]{appendix:efficiency}
We report fine-tuning and inference costs in \Cref{tab:computational_efficiency}, measured on CORINE in BF16 with effective batch sizes of 32 and 1, respectively. Since ChannelViT exceeds GPU memory at a fine-tuning batch size of 32, its step time is measured using a microbatch size of 4 with eight gradient accumulation steps per optimizer update.

Compared with ChannelViT, LESSViT achieves
approximately a 29-fold reduction in fine-tuning
step time, a 7-fold reduction in inference latency,
and an 11-fold reduction in inference memory,
while requiring approximately 2.6 times less
estimated peak fine-tuning memory. It remains more expensive to train than channel-collapsing methods because it retains the explicit $N\times C$ spatial--spectral token grid. Our efficiency advantage is therefore relative to full spatial--spectral attention. Profiling shows that LESS Attention accounts for only 1.17 GB, approximately 1.3\% of peak training memory, whereas MLP activations account for approximately 43\%. The remaining memory overhead thus primarily reflects processing the explicit token grid rather than constructing a quadratic attention matrix. Pretraining costs are reported in \Cref{appendix:pretrain}.
\begin{table}[h]
\centering
\small
\renewcommand{\arraystretch}{1.15}
\setlength{\tabcolsep}{5pt}
\caption{\textbf{Computational efficiency.} Encoder parameter counts and fine-tuning and inference costs.}
\label{tab:computational_efficiency}
\resizebox{0.9\textwidth}{!}{%
\begin{tabular}{lccccc}
\toprule
& Encoder & \multicolumn{2}{c}{Fine-tuning} & \multicolumn{2}{c}{Inference} \\
\cmidrule(lr){3-4}\cmidrule(lr){5-6}
Model & params (M) & Step time (s) & Peak memory (GB) & Latency (ms) & Peak memory (GB) \\
\midrule
SpectralViT-B & 87.5 & 0.39 & 14.9 & 7.1 & 0.64 \\
HyperSigma-B & 203.0 & 0.17 & 12.2 & 32.0 & 1.31 \\
DOFA-B & 111.8 & 0.035 & 1.93 & 13.3 & 0.87 \\
HyperFree-B & 151.5 & 0.18 & 2.25 & 52.8 & 0.94 \\
ChannelViT-B$^{\ast}$ & 85.5 & 21.0 & $\sim$230 & 221.4 & 7.73 \\
\rowcolor[HTML]{ECF4FF}
LESSViT-B & 83.2 & 0.73 & 87.2 & 30.5 & 0.67 \\
\bottomrule
\end{tabular}%
}
\par\vspace{4pt}
\begin{minipage}{\textwidth}
\footnotesize
$^{\ast}$ChannelViT exceeds GPU memory at batch size 32; the approximately 230 GB memory requirement is estimated from the OOM error. Its step time is measured using a microbatch size of 4 with eight gradient accumulation steps per optimizer update.
\end{minipage}
\end{table}

\subsection{Additional Qualitative Results}
\begin{figure}[h]
    \centering
    \includegraphics[width=0.9\linewidth]{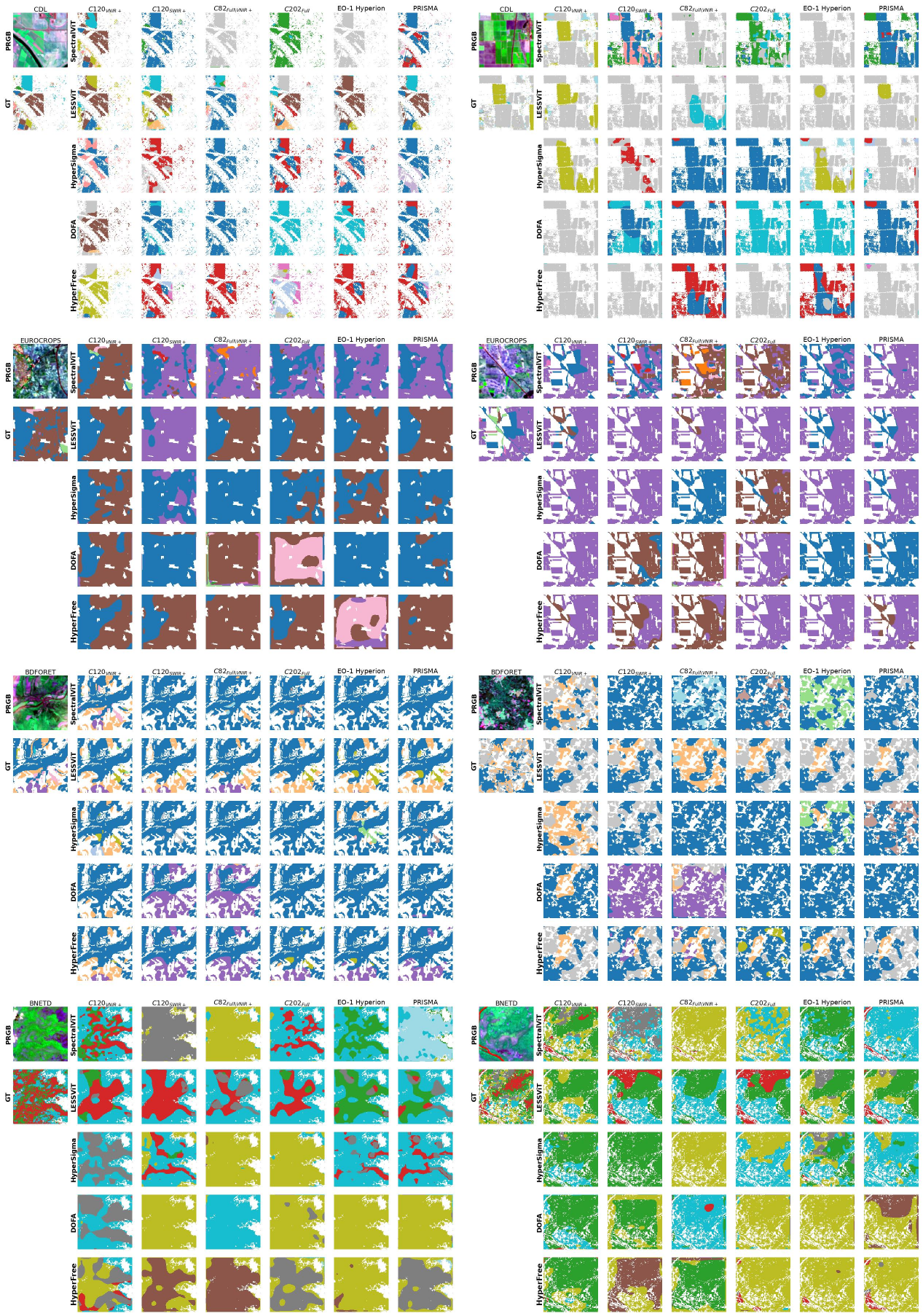}
    \caption{\textbf{Additional qualitative results for segmentation tasks.} We visualize the segmentation maps generated by SpectralViT, LESSViT, HyperSigma, and DOFA on different tasks and channel configurations.}
    \label{fig:all_vis}
\end{figure}

We provide additional qualitative results for segmentation tasks in \Cref{fig:all_vis}.

%%%%%%%%%%%%%%%%%%%%%%%%%%%%%%%%%%%%%%%%%%%%%%%%%%%%%%%%%%%%

% \newpage
% \input{checklist.tex}

\end{document}